 \documentclass[final,5p,times,twocolumn,authoryear]{elsarticle}

\usepackage{amssymb}
\usepackage{lipsum}
\usepackage{amsmath}
\usepackage{booktabs}
\usepackage{tabularx}
\usepackage{longtable}
\usepackage{subfigure}
\usepackage{comment}
\usepackage{url}
\usepackage{xurl}
\usepackage{xcolor}
\usepackage{hyperref}

\journal{Natural Language Processing Journal}

\begin{document}

\begin{frontmatter}



\title{Opportunities and Challenges of Natural Language Processing for Low-Resource \textsc{Senegalese} Languages in Social Science Research}


    
\affiliation[1]{
  organization={Polytechnic School (ESP)},
  city={Dakar},
  country={Senegal}
}

\affiliation[2]{
  organization={Gaston Berger University (UGB)},
  city={Saint Louis},
  country={Senegal}
}

\affiliation[3]{
  organization={Federal Institute of Technology Lausanne (EPFL)},
  city={Lausanne},
  country={Switzerland}
}

\author[1]{Derguene Mbaye}
\affiliation{ead={derguenembaye@esp.sn}}

\author[2]{Tatiana D. P. Mbengue}

\author[1]{Madoune R. Seye}

\author[1]{Moussa Diallo}

\author[1]{Mamadou L. Ndiaye}

\author[2]{Dimitri S. Adjanohoun}

\author[2]{Djiby Sow}

\author[2]{Cheikh S. Wade}

\author[3]{Jean-Claude B. Munyaka}

\author[3]{Jerome Chenal}

\begin{abstract}
Natural Language Processing (NLP) is rapidly transforming research methodologies across disciplines, yet African languages remain largely underrepresented in this technological shift. This paper provides the first comprehensive overview of NLP progress and challenges for the six national languages officially recognized by the Senegalese Constitution: \texttt{Wolof}, \texttt{Pulaar}, \texttt{Sérère}, \texttt{Diola}, \texttt{Mandingue}, and \texttt{Soninké}. We synthesize linguistic, sociotechnical, and infrastructural factors that shape their digital readiness and identify gaps in data, tools, and benchmarks. Building on existing initiatives and research works, we analyze ongoing efforts in various tasks, covering both text and speech modalities. We also provide a centralized GitHub repository that compiles publicly accessible resources for a range of NLP tasks across these languages, designed to facilitate collaboration and reproducibility. A special focus is devoted to the application of NLP to the \texttt{social sciences}, where multilingual transcription, translation, and retrieval pipelines can significantly enhance the efficiency and inclusiveness of field research. The paper concludes by outlining a roadmap toward sustainable, community-centered NLP ecosystems for Senegalese languages, emphasizing ethical data governance, open resources, and interdisciplinary collaboration.
\end{abstract}


\end{frontmatter}




\section{Introduction}
Natural Language Processing (NLP) has emerged as a transformative field within artificial intelligence (AI), enabling machines to process and understand human language at scale. In recent years, its applications have profoundly influenced research across disciplines, from computational linguistics and digital humanities to sociology and political science. However, the vast majority of NLP advances have been concentrated on a small set of \texttt{high-resource} languages, leaving most African (\texttt{low-resource}) languages under-represented in both datasets and algorithmic development \citep{nekoto-etal-2020-participatory}. However, the term “low resource” can cover broader dimensions and is not limited to language alone: it can also refer to \texttt{domains} or \texttt{tasks} for which little data is available, even when the language in question is rich in resources. This is particularly evident in \citep{nlplrlsurvey}, where the concept of "low resource" is defined according to three distinct aspects: 

\begin{itemize}
    \item The availability of \textbf{task-specific} annotations ;
    \item The existence of \textbf{unannotated} texts in the language ;
    \item The presence of \textbf{auxiliary} data.
\end{itemize}

\begin{figure}
\centering
\resizebox*{9cm}{!}{\includegraphics{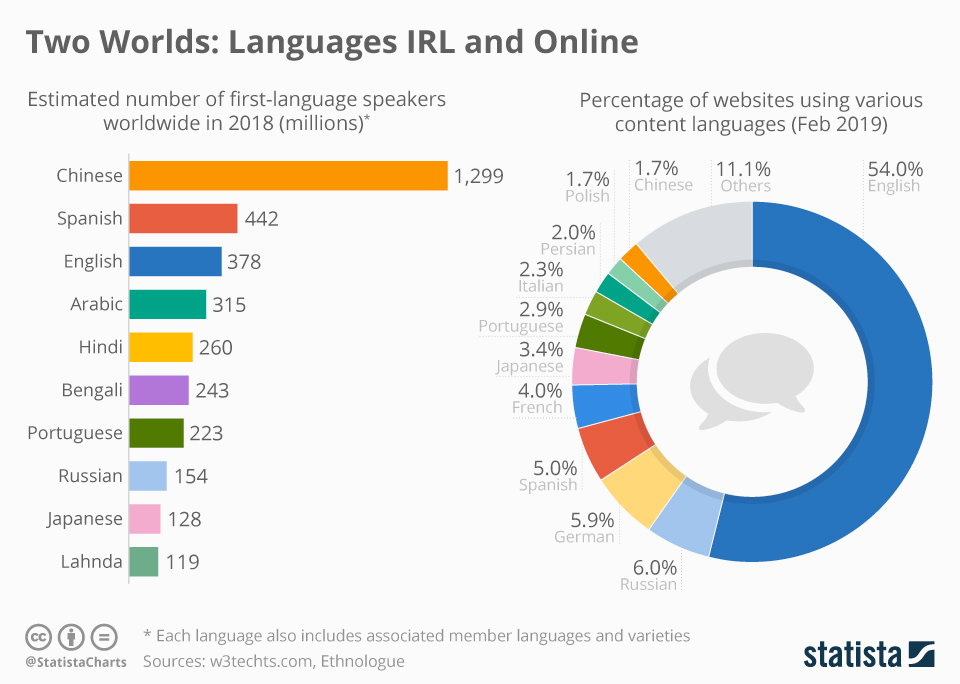}}
\caption{Contrast between the world's most spoken languages and their representation online.} \label{lang-contrast}
\end{figure}

As illustrated in \citep{adebara-abdul-mageed-2022-towards}, the majority of African languages fit this definition, which complicates the work of researchers, and contributes to their under-representation in NLP research \citep{51306}. Evaluation results from the \texttt{SAHARA} benchmark \citep{adebara-etal-2025}, which assesses 517 African languages across multiple NLP tasks, reveal a pronounced performance gap: English consistently ranks among the highest-performing languages, whereas many widely spoken African languages, such as \textbf{Pulaar} (Fula or Fulfulde), \textbf{Wolof}, Hausa, Oromo, and Kinyarwanda, systematically underperform across reasoning, generation, and classification tasks. These widening linguistic inequities in AI capabilities is particularly alarming as a new type of digital divide\footnote{Traditionally been framed in terms of access to internet connectivity.} is emerging, which now concerns \texttt{the extent to which languages are represented and processed by AI systems} \citep{olofsson}. 

Senegal, a multilingual nation with over twenty languages, constitutionally recognizes six national languages\footnote{Presidential Decree No 71-566 of May 21, 1971.}: \textbf{Wolof}, \textbf{Pulaar}, \textbf{Sérère}, \textbf{Diola}, \textbf{Mandingue}, and \textbf{Soninké}, as central to its cultural and civic identity. Although a minority, there are also populations of Arabic (Afro-asiatic) speakers, including those who speak \texttt{Hassaniyya} (Mauritanian dialect of Arabic), as well as \texttt{Levantine} and \texttt{Moroccan dialects}. \texttt{Portuguese Creole} is also spoken in some parts of Casamance, and in Dakar by immigrant and migrant populations from the Cape Verde islands and Casamance respectively \citep{langafrica}. However, \texttt{French} remains the dominant European language in Senegal, being recognized as the official language and the one used in education \citep{langafrica}. In this article, we will only focus on the 06 main local languages recognized as \texttt{national} languages and leave the other ones for future research. Despite their societal importance and widespread use in daily communication and media, these languages remain largely excluded from the digital and scientific landscape of NLP. This gap poses a dual challenge: the risk of technological marginalization of major linguistic communities, and the missed opportunity to harness NLP for advancing locally grounded research, especially in the social sciences.

Social science research in Senegal relies heavily on qualitative methods, including interviews, focus groups, and ethnographic recordings \citep{faye2025, dime2019} often multilingual and resource-intensive to transcribe, translate, and analyze. The integration of NLP pipelines into these workflows could dramatically improve efficiency, accessibility, and analytic depth. Yet, realizing this potential requires robust linguistic resources, interoperable tools, and sustainable community infrastructures.  Linguistic inequity in AI constitutes a structural issue rather than a marginal fairness concern, as it directly shapes access to reliable information, the ability to challenge decisions, and \texttt{meaningful participation in democratic processes} \citep{olofsson}. 

The growing body of NLP research on African languages has produced several broad surveys and resource compilations, yet none focuses exclusively on Senegal’s constitutionally recognized national languages. At the continental scale, \citet{belay2026riseafricanlpsurvey} surveyed over 1,900 papers across 184 African languages and 44 NLP task categories; their bibliometric analysis reveals that \textbf{Wolof}, with only 32 identified papers, is the sole Senegalese language with \textbf{Pulaar (Fulfulde)} among the 17 most-studied African languages, while \textbf{Sérère}, \textbf{Diola}, \textbf{Mandingue}, and \textbf{Soninké} are essentially invisible in the continental corpus. Regional surveys such as \citet{tonja2023} on the Ethiopian NLP ecosystem demonstrate the value of country-level depth, but no equivalent work exists for Senegal. Broader resource catalogues (e.g., \citet{adebara-abdul-mageed-2022-towards}) provide language-level inventories without the task-by-task, language-by-language analysis or the social science framing that characterize the present paper. To our knowledge, this survey is therefore the first to \textbf{(a)} cover all six Senegalese national languages simultaneously, \textbf{(b)} map NLP progress across the full task spectrum; from language identification and parsing to speech synthesis and spoken dialogue systems; and \textbf{(c)} explicitly situate these technical findings within the needs of social science field research. In addition, we introduce a centralized and openly accessible repository on GitHub\footnote{\url{https://github.com/DerXter/State-of-NLP-Research-in-Senegal}} that compiles existing datasets, benchmarks, and tools available for these languages. The repository is designed as a living resource to be periodically expanded through community contributions. Our objective is to map existing efforts, identify critical research gaps, and encourage the development of sustainable, inclusive NLP research for Senegal’s national languages. 

\section{Survey Methodology}
\label{sec:methodology}
\begin{figure*}[!htbp]
\centering
\includegraphics[width=1\textwidth]{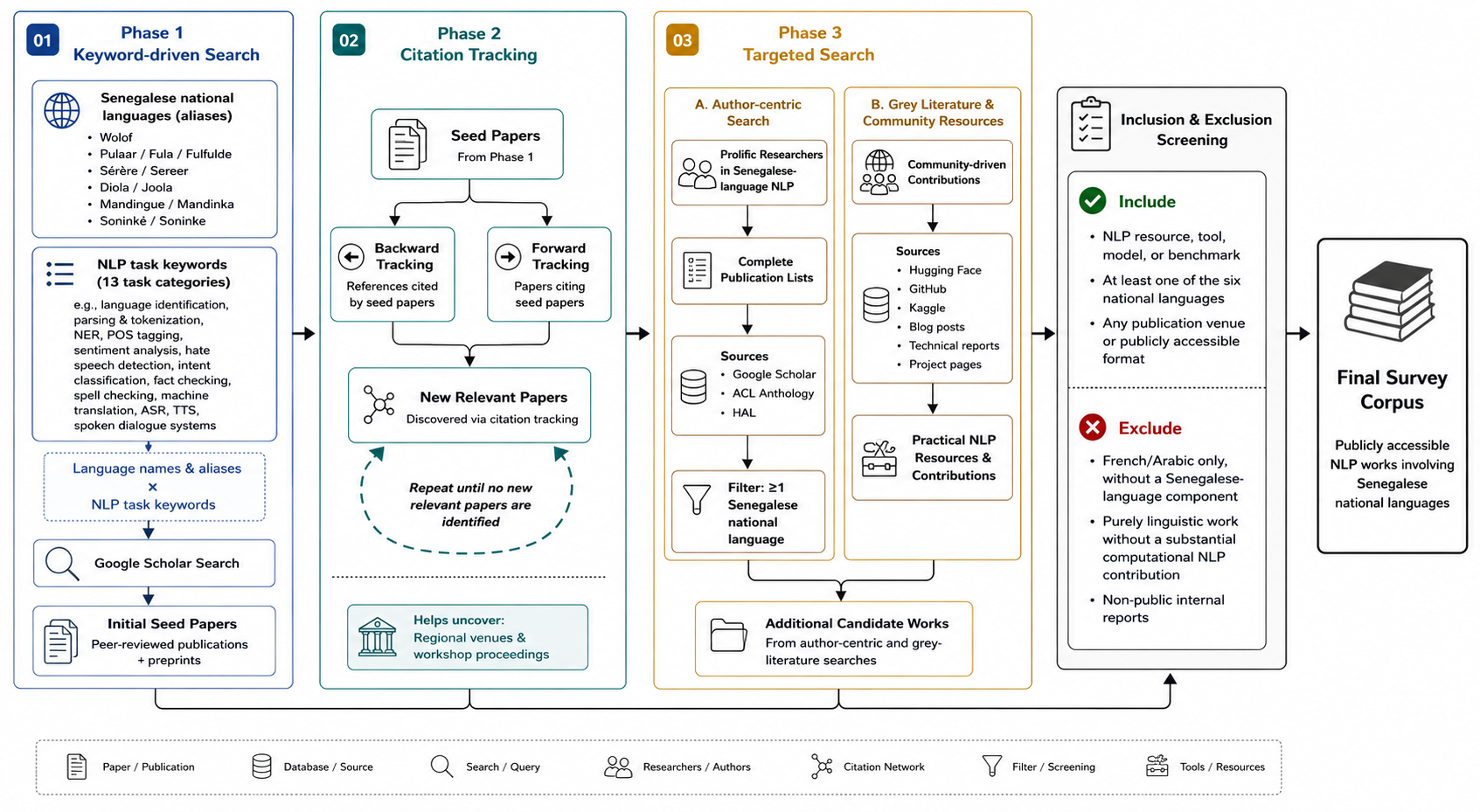}
\caption{Survey paper collection process. The three-phase pipeline combines keyword-driven Google Scholar queries, forward and backward citation tracking, and author-centric plus grey-literature targeting. The resulting raw pool is then deduplicated and filtered against explicit inclusion and exclusion criteria.}
\label{fig:methodology}
\end{figure*}

The corpus of works reviewed in this paper was assembled through a three-phase search process designed to ensure broad coverage of a field that is dispersed across traditional publishing venues and informal community channels, as illustrated in Figure~\ref{fig:methodology}.

\textbf{Keyword-driven search.} We queried \texttt{Google Scholar}\footnote{https://scholar.google.com/} using combinations of each of the six national language names and their principal aliases (e.g., Wolof, Pulaar/Fula/Fulfulde, Sérère/Sereer, Diola/Joola, Mandingue/Mandinka, Soninké/Soninke) paired with keywords for each NLP task category covered in Section~\ref{sec:effort}. This yielded an initial seed set of peer-reviewed publications and preprints.

\textbf{Citation tracking.} For each seed paper identified in the first phase, we performed both forward citation tracking (papers that cite the seed) and backward reference tracking (the seed’s own reference list), iterating until no new relevant papers were discovered. This step proved particularly important for surfacing works published in regional venues or workshop proceedings that do not rank prominently in keyword searches.

\textbf{Author-centric and grey-literature targeting.} We identified prolific researchers active in Senegalese-language NLP and retrieved their complete publication lists on Google Scholar, ACL Anthology, and the French open-archive server \texttt{HAL}\footnote{\url{https://hal.science/?lang=en}}, filtering for papers involving at least one national language. In addition, we tracked community-driven contributions that do not appear in traditional bibliographies, including blog posts, technical reports, and project pages on platforms such as \texttt{Hugging Face}\footnote{\url{https://huggingface.co}}, \texttt{GitHub}\footnote{\url{https://github.com}}, and \texttt{Kaggle}\footnote{\url{https://www.kaggle.com}}, which represent a substantial and often overlooked fraction of the practical resource base for low-resource languages.

\textbf{Inclusion and exclusion criteria.} We included any work producing or evaluating an NLP resource, tool, model, or benchmark for at least one of the six national languages, regardless of publication venue or format. We excluded papers dealing exclusively with French or Arabic as pivot languages without a Senegalese-language component, works with no substantial NLP contribution (e.g., purely linguistic field documentation without computational tools), and unpublished internal reports that are not publicly accessible.

\section{Sociolinguistic and Linguistic Features}
\begin{figure}
\centering
\resizebox*{9cm}{!}{\includegraphics{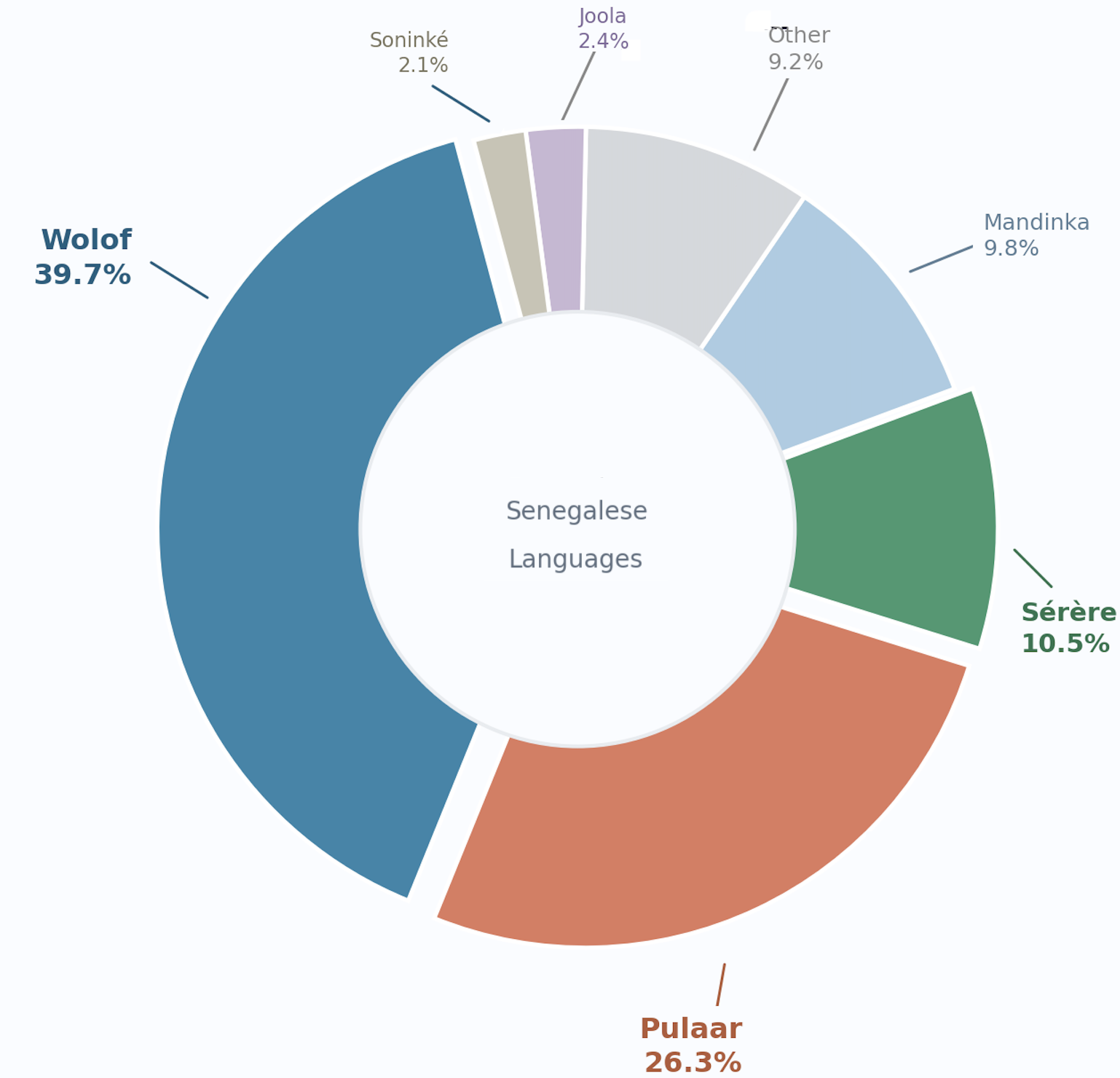}}
\caption{Proportion of speakers of the main local languages in terms of percentages \citep{leclerc2015}.} \label{fig:langscape}
\end{figure}

The Senegalese linguistic landscape is characterized by considerable diversity, with approximately 30 national languages, most of which belong to the \texttt{Niger–Congo language family} \citep{leclerc2015}. Among these, 06 languages: \textbf{Wolof}, \textbf{Pulaar (Fula)}, \textbf{Sereer}, \textbf{Diola (Joola)}, \textbf{Mandinka (Malinké)}, and \textbf{Soninke}; occupy a particularly prominent position. From a demographic perspective, only \texttt{Wolof}, \texttt{Pulaar}, \texttt{Serer}, and \texttt{Mandinka} are spoken by more than one million speakers\footnote{Senegal's population is approximately 18.5 million people, with recent estimates \citep{worldbank}.} as illustrated in Figure \ref{fig:langscape}. 

These languages have officially standardized orthographies recognized and enforced by the Senegalese state, making them, in principle, suitable for integration into the educational system. As early as 1971, Presidential Decree No. \texttt{71-566 of May 21, 1971} formally recognized these languages as \textbf{“national languages”}, while \textbf{French} retains the status of the country’s “official language”, being therefore \texttt{the sole language of education and administration} \citep{langes-etat}. The integration of national languages in the educational system remains thus very weak, and further work remains necessary, particularly with regard to pedagogical standardization and the production of appropriate teaching materials \citep{leclerc2015}. Attempts to introduce national languages into the school system have indeed taken place, notably through the experimental classes of 1977–1979 and the recommendations of the 1981 and 1984 reforms, but they have all been thwarted by a lack of political will, financial resources, and coherent planning. Wolof, although dominant in daily life, the media, and election campaigns, thus remains marginalized in formal education, perpetuating a deep divide between the country’s linguistic reality and its official education system \citep{langes-etat}. However, other more recent projects to integrate local languages were subsequently initiated. Launched in 2011 by the Institut de la Francophonie pour l’éducation et la formation (\texttt{IFEF}), the \texttt{ELAN} (École et Langues Nationales) project aimed to establish bilingual education in French and the national language in primary schools across several French-speaking African countries, including Senegal \citep{ndione:hal-04209973}. By 2017, the program already included 2,500 schools and more than 225,000 students across 12 countries, with 35 national languages used as languages of instruction alongside French. Integrated into the Senegalese education system from its inception, this project provided the pedagogical foundation and the bilingual and multilingual materials essential to the development of the Harmonized Model for Bilingual Education in Senegal \texttt{(MOHEBS)} \citep{afd-ifef-elan2-2026}. This national model is now undergoing a major strategic expansion covering 13 of the country’s 16 educational regions for six standardized local languages, and has become a benchmark for other countries in the subregion \citep{gates-edu}. The 2025–2026 Implementation Status and Funding Needs Report, indicates that 7,325 schools are already covered \citep{rapport-mohebs} out of 11,181 elementary schools in Senegal\footnote{Count conducted by the ANSD in the 2024 Report on the Economic and Social Situation of Senegal \citep{sesn2024}. }, bringing the total coverage rate to 65.51\%. Despite these promising results, the Ministry of National Education (MEN) notes that major challenges remain, such as the availability of textbooks, strengthening teacher training,  as well as program management and coordination, among others \citep{men2025}. Nevertheless, they are exploring ways to use AI and NLP in collaboration with local stakeholders to overcome these limitations \citep{idinsight2026yidan, idinsight2026bilingual}, which underscores the importance of the present work. These challenges hinder the expansion of the \texttt{MOHEBS} program, which is, moreover, currently limited to elementary education, excluding middle schools, high schools, and universities\footnote{There are, however, local university programs focused on the study of these languages.}. Furthermore, because the project is so recent, entire generations have received their education exclusively in French, creating a general climate of limited proficiency in the national languages. Consequently, few speakers adhere to their spelling rules, which makes the available texts; particularly on social media; highly inconsistent and difficult to analyze. Many of these languages, especially Pulaar and Mandinka, are also characterized by substantial dialectal variation, which poses additional challenges for linguistic standardization and computational processing.

\subsection{Language Digraphs}
Although the \texttt{Latin script} is the most widely used, some languages such as Wolof, Pulaar, Mandinka and Soninké also have an Arabic-based alphabet called \texttt{Ajami} \citep{ouzerrout-saadallah}. It results from the early Islamization of the major Muslim ethnic groups in the country, and remains an important means of written communication among people who are illiterate in French and who have attended Quranic schools \citep{ajami}. This phenomenon has thus created two distinct worlds that do not converge, and each writing system has its own applications. While the Ajami script is mainly used in religious contexts and traditional medicine, the Latin script is widely used in the digital world, particularly for the localization of online platforms \citep{nguer2020digraphs}. 

The history of the Ajami script has been explored in \citep{le-etal-2025-best}, as well as its use and its modern writings. Authors analyzed the challenges and prospects of these systems from the perspective of language preservation and highlighted the potential of this script to represent an important instrument for literacy and digital inclusion in sub-Saharan Africa. Therefore, considerable efforts have been made in order to promote its scriptural rehabilitation through \texttt{transliteration}\footnote{Process of representing a word, phrase, or text in a different script or writing system.}. Challenges in Latin-Ajami transliteration have been explored in \citep{nguer2020digraphs} with a purpose of involving people using the Arabic alphabet within a collaborative dictionnary project. The creation of \texttt{Latin2Ajami}: a transliteration algorithm from latin towards Ajami, has been studied in \citep{fall2020digraphies} with an approach based on a correspondence table, whose data comes from an external editable file. The \texttt{AjamiXTranslit} project went further by offering a collaborative data collection platform for native speakers and a publicly available multilingual corpus of Latin–Ajami text pairs along with annotated manuscripts \citep{ouzerrout-saadallah}. The authors also introduced automatic transliteration and optical character recognition (OCR) models adapted to the graphic diversity of Ajami. Although small in size (the largest of the Senegalese languages < 70 rows), this is the corpus with the widest coverage of Senegalese languages (Wolof, Pulaar, and Soninké) in Ajami script.

\subsection{Language Overview}
\begin{figure}
\centering
\resizebox*{9cm}{!}{\includegraphics{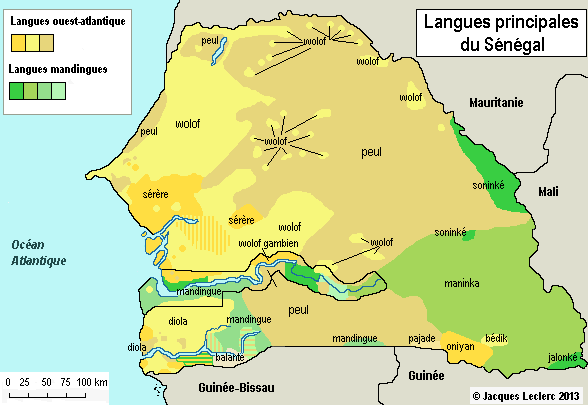}}
\caption{Main Senegalese languages and their locations in the country \citep{leclerc2015}.} \label{fig:langloc}
\end{figure}

While most of the senegalese languages belong to the Niger–Congo language family, the majority of them belong either to the \texttt{West Atlantic group} or \texttt{the Mandinka group} \citep{langafrica}. They encompass a wide range of phonological, morphological, and orthographic systems and are more localized in different areas in Senegal, as illustrated in the Figure \ref{fig:langloc}. This section outlines the main sociolinguistic profiles of these dominant languages in Senegal as well as linguistic characteristics relevant to NLP development.

\subsubsection{Wolof}
Wolof serves as the dominant lingua franca of Senegal, spoken by more than 80\% of the population \texttt{either as a first or second language}. It belongs to the Atlantic branch of the Niger–Congo family and exhibits rich morphophonemic alternations, vowel harmony, and extensive use of nasalization. Orthographic variation remains a challenge, especially around diacritics such as \textit{ñ} and \textit{$\eta$} ("ng"). Due to its sociolinguistic reach and early digitalization efforts, Wolof has become the most resourced Senegalese language in NLP research to date as illustrated in Tab \ref{tab:resources}. There are two main geographical varieties of Wolof: \texttt{one spoken in Senegal, and the other in Gambia} \citep{ethnologue-2019}. Although Wolof speakers understand each other, Senegalese Wolof and Gambian Wolof are considered two distinct languages, each with its own ISO 639-3 language code: \texttt{WOL} for Senegalese Wolof and \texttt{WOF} for the Gambian one \citep{gauthier-etal-2016}. Wolof has also been adapted to the Latin alphabet, despite having a long written tradition using the Arabic alphabet in the form of Ajami (or Wolofal) \citep{le-etal-2025-best}. Despite showing remarkably little variation accross dialects, the major contemporary Wolof dialectal divide is between \texttt{rural} and \texttt{urban} varieties. The latter has extensively borrowed from French as the result of language contact \citep{langafrica} which led to the \texttt{code-switching} phenomenon, making it more challenging to process using computer tools \citep{cetinoglu-etal-2016-challenges}.

\subsubsection{Pulaar / Fula}
Pulaar, also known as \textbf{Fula}\footnote{This variant is usually used in NLP research papers.} or \texttt{Fulfulde}, is part of the Atlantic family and is spoken across West and Central Africa. In Senegal, it is mainly concentrated in the \texttt{Fouta Toro} region (cf Fig \ref{fig:langloc}) whose dialect being the most dominant one across several mutually intelligible varieties \citep{langafrica}. Pulaar exhibits complex noun class systems, consisting of more than 20 classes in some dialects of Pulaar, a large set of verbal extensions, and morphologically conditioned consonant mutation.

Other dialects spoken in Senegal include the \texttt{Casamance} dialects such as \texttt{Fulakunda} and \texttt{Fulaadu}, as well as the \texttt{Fuuta Jalon} dialect spoken by the substantial population of \texttt{Guineans} living in Senegal  \citep{langafrica}. Its transnational presence makes it an ideal candidate for \texttt{regional NLP initiatives}, although orthographic harmonization across borders remains incomplete.

\subsubsection{Sérère (Sereer)}
Sérère, belongs to the northern branch of the Atlantic family of the Niger-Congo phylum, which makes it related to \texttt{Wolof} and especially \texttt{Fula} \citep{serere}. The different varieties of Serer are spoken by more than 1 million people in Senegal and Gambia (cf Fig \ref{fig:langscape}), but it is important to note that \texttt{this term also refers to populations in eastern Senegal who are culturally similar but speak Cangin languages}\footnote{\url{https://en.wikipedia.org/wiki/Cangin_languages}}. Recognized as one of the national languages of Senegal, it has an official writing system based on the variety known as \texttt{Seereer Siin}, meaning "Sereer from the Sine region" between the \textbf{Petite Côte} south of Dakar and Gambia, which has become a kind of “standard” Sereer \citep{serere}. Despite its significant speaker base, it remains severely underrepresented in digital corpora. Documentation efforts are growing through community and academic collaborations, but resources for NLP applications remain still minimal (cf Tab \ref{tab:resources}).

\subsubsection{Diola (Joola)}
The Joola group comprises several dialects such as \texttt{Joola Foñi} (dominant dialect) or \texttt{Kujamaat Joola}, as well as \texttt{Boulouf}, \texttt{Fogny} and \texttt{Kasa} \citep{leclerc2015}. It belongs to the \texttt{Bak} branch of Atlantic \citep{langafrica} and is primarily spoken in the Casamance region (cf Fig \ref{fig:langloc}). It shares certain typological features with \texttt{Wolof}, \texttt{Pulaar} and \texttt{Sereer} (Pulaar's closest language), and are unusual among Niger-Congo languages in that \texttt{they are not tonal} \citep{langafrica}. Joola's morphosyntactic diversity make it particularly challenging for corpus alignment, and limited orthographic standardization and dialectal fragmentation also contribute to data scarcity. It is rarely used in writing but rather on local radio stations, limiting its use to oral communication in everyday life \citep{joola}. However, ongoing linguistic documentation projects are beginning to fill the gap.

\subsubsection{Mandingue (Mandinka)}
A member of the \texttt{Mande} language family, \texttt{Mandingue} is widely spoken in Senegal’s southern regions and across neighboring countries (cf Fig \ref{fig:langloc}). About half of Mandinka speakers live in \texttt{Gambia}, where Mandinka is the dominant language nationwide \citep{mandinka}. With \texttt{Wolof}, \texttt{Pulaar} and \texttt{Sereer}, they represent the dominant languages in Senegal in terms of numbers of speakers (cf Fig \ref{fig:langscape}). The most salient features of Mandinka are very similar to those of other Manding languages (particularly \textbf{Bambara}) such as \textbf{(1)} a \texttt{tonal} system based on the opposition between high and low tones, \textbf{(2)} the virtual absence of syllables ending in a consonant, with the exception of syllables ending in the nasal $\eta$, \textbf{(3)} a very limited range of morphological inflection, \textbf{(4)} the absence of grammatical gender and, \textbf{(5)} an extremely rigid word order \citep{mandinka}. Its relatively stable morphology and regional presence, offer potential for \texttt{transfer learning}\footnote{A technique in machine learning (ML) in which knowledge learned from a task/language, is re-used in order to boost performance on a related task/language.} from related Mande languages such as \texttt{Bambara}, in which substantial NLP resources are being developed \citep{tapo}. 

\subsubsection{Soninké}
Soninké, one of the oldest written languages in West Africa, belongs to the \texttt{Mande} family and retains strong oral traditions. It plays a key role in historical, cultural, and economic exchanges across \texttt{Mali}, \texttt{Mauritania}, and \texttt{Senegal} \citep{soninke}. Soninké is a \textbf{tonal language}, in which each syllable is characterized by a musical pitch, either high (\texttt{´}) or low (\texttt{`}), and represents somewhat of an outlier, as it has little mutual intelligibility with other Mande languages \citep{langafrica}. However, the differences between the various dialectal varieties \texttt{within Soninke} are relatively minor and do not hinder mutual intelligibility \citep{soninke}. Moreover, there is no standard form of Soninke, nor is there a dialectal variety of this language that is recognized as more dominant than the others \citep{soninke}. 
While limited digital resources exist, Soninké’s cross-border usage and structured morphology make it a promising target for multilingual modeling.

\section{Low-Resource Context and Data Availability}\label{sec:lrcontext}

\begin{figure*}[!htbp]
\centering
\includegraphics[width=1\textwidth]{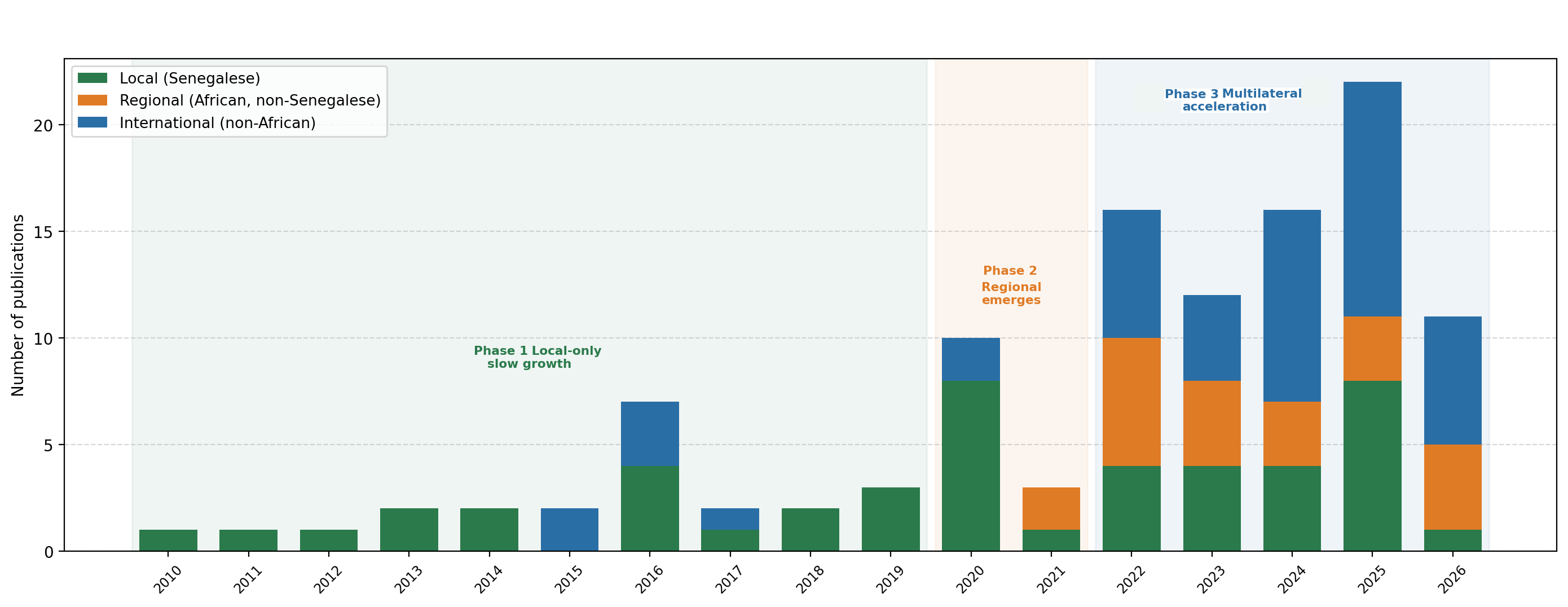}
\caption{Publication growth for Senegalese NLP research from 2010 to 2026 (peer-reviewed papers only), up to the survey cutoff date.}
\label{fig:timeline}
\end{figure*}

Senegal’s NLP ecosystem remains in an early but dynamic stage of development as shown in Figure~\ref{fig:timeline}. Three phases are visible: an initial local-only period (2010--2019), the emergence of regional African collaborations around 2021, and a multilateral acceleration from 2022, onward driven by the inclusion of Wolof and Pulaar in large-scale multilingual benchmarks (cf Section~\ref{sec:effort}). 

\begin{figure}[!htbp]
\centering
\includegraphics[width=\linewidth]{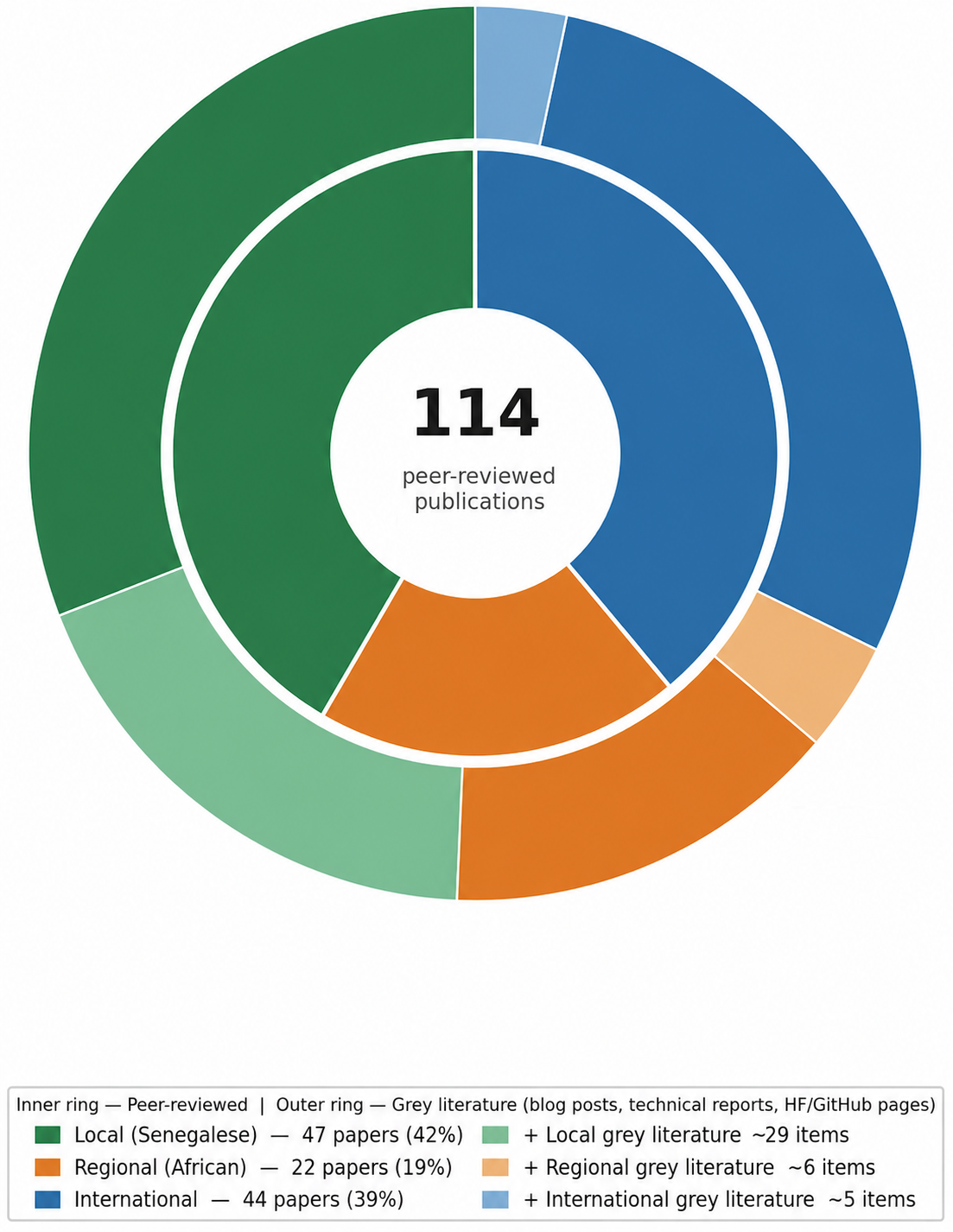}
\caption{Overall origin distribution of Senegalese NLP publications (beyond peer-reviewed papers).}
\label{fig:donut}
\end{figure}

While data scarcity and resource fragmentation continue to constrain progress, significant institutional and community-driven initiatives are emerging to bridge this gap. The establishment of the \textbf{national AI strategy} has marked a pivotal step toward building national capacity in artificial intelligence and data governance \citep{aistrategy}. From an \textbf{academic} standpoint, several institutions have contributed to the early stages of resource development and linguistic documentation, supporting foundational research and training in machine learning and natural language processing \citep{sennlp}. \textbf{Community-driven movements} have also been instrumental. \textbf{GalsenAI} and \textbf{Masakhane} have spearheaded open, collaborative data collection and multilingual modeling across African languages \citep{grallet2025}. These networks have facilitated the creation of various datasets, tools and publications contributing to the first generation of open benchmarks for low-resource NLP. Beyond peer-reviewed papers, Figure~\ref{fig:donut} shows the total number of publications covered in this survey (inner ring) split by research origin: \texttt{Local} (Senegalese, 42\%), \texttt{Regional }(African non-Senegalese, 19\%), and \texttt{International} (non-African, 39\%). The outer ring adds the approximate count of \texttt{grey-literature contributions} (blog posts, technical reports, Hugging Face and GitHub project pages), which are disproportionately local in origin, reflecting the community-driven nature of practical resource development. Moreover, \textbf{Private companies} and \textbf{start-ups} also contribute to the development of NLP resources \citep{HENG2022102454}, rounding out this overview. 

Table~\ref{tab:resources} summarizes some of the publicly known datasets, corpora, and tools relevant to Senegalese national languages, highlighting \texttt{the current imbalance between Wolof and other local languages}. A separate and more exhaustive GitHub document\footnote{\url{https://github.com/DerXter/State-of-NLP-Research-in-Senegal/blob/main/Datasets.md}} which includes additional \texttt{data sources}, has also been set up to facilitate the tracking of datasets continuously produced in these languages.

\begin{table}[!htbp]
\centering
\caption{\texttt{Excerpt} of publicly available \texttt{datasets} and \texttt{tools} for Senegalese \textbf{national} languages covering \textcolor{red}{Machine Translation}, \textcolor{blue}{Token Classification}, \textcolor{purple}{QA/Instruct/Pre-Training}, \textcolor{orange}{Sentence Classification}, \textcolor{brown}{Automatic Speech Recognition}, \textcolor{violet}{Speech Synthesis}, \textcolor{teal}{Keyword Spotting}, \textcolor{magenta}{Language Identification}, \textcolor{cyan}{Morphological features tagging \& alignment evaluation}, \textcolor{green}{Masked Language Modeling}. Every entry is hyperlinked directly to its source repository or download page. A continuously updated catalogue of resources across all NLP tasks and languages is maintained in the companion repository introduced in this work.}

\label{tab:resources}
\begin{tabularx}{\linewidth}{@{}lXXXX@{}}
\toprule
\textbf{Language} & \textbf{Text corpora} & \textbf{Speech corpora} & \textbf{Existing tools} \\
\midrule
Wolof & \textcolor{red}{\href{https://opus.nlpl.eu/}{OPUS}, \href{https://github.com/facebookresearch/flores/blob/main/flores200/README.md}{FLORES-200}, \href{https://github.com/MicrosoftTranslator/NTREX}{NTREX}, \href{https://catalog.ldc.upenn.edu/LDC2022T03}{LORELEI}, \href{https://github.com/masakhane-io/lafand-mt/}{MAFAND-MT}, \href{https://huggingface.co/datasets/google/smol}{SMOL}, \href{https://github.com/google-research/google-research/tree/master/madlad_400}{MADLAD-400}, \href{https://huggingface.co/datasets/HuggingFaceFW/finetranslations}{FineTranslations}}, \textcolor{blue}{\href{https://github.com/masakhane-io/masakhane-ner}{MasakhaNER}, \href{https://github.com/masakhane-io/masakhane-pos}{MasakhaPOS} \href{https://github.com/UniversalDependencies/UD_Wolof-WTB}{Universal Dependencies}}, \textcolor{purple}{\href{https://huggingface.co/datasets/masakhane/afriqa}{AfriQA}, \href{https://huggingface.co/datasets/CohereForAI/aya_collection_language_split/viewer/wolof}{AYA}, \href{https://github.com/masakhane-io/chatbots-african-languages/}{AfriWOZ-1.0}, \href{https://github.com/facebookresearch/belebele}{Belebele}, \href{https://huggingface.co/datasets/soynade-research/FineWeb2-HQ-50k-Wolof}{FineWeb-Wolof-50k (Synthetic Corpus)}}, \textcolor{orange}{\href{https://zenodo.org/records/14497425}{AWOFRO}, \href{https://www.kaggle.com/datasets/abdoukarimkandji/wolbanking77}{WolBanking77}, \href{https://github.com/masakhane-io/masakhane-nlu}{Masakhane-NLU}}, \href{https://huggingface.co/datasets/TutlaytAI/AjamiXTranslit}{AjamiXTranslit} & \textcolor{brown}{\href{https://zindi.africa/competitions/ai4d-baamtu-datamation-automatic-speech-recognition-in-wolof/data}{AI4D-Urban}, \href{https://github.com/getalp/ALFFA_PUBLIC/tree/master/ASR/WOLOF}{ALFFA}, \href{https://www.kaggle.com/datasets/abdoukarimkandji/wolbanking77}{WolBanking77}, \href{https://github.com/Waxal-Multilingual/speech-data}{Waxal}, \href{https://huggingface.co/datasets/google/fleurs/viewer/wo_sn}{Fleurs}, \href{https://github.com/gauthelo/kallaama-speech-dataset}{Kallaama}, \href{https://huggingface.co/datasets/galsenai/WaxalNLP}{WaxalNLP}}, \textcolor{violet}{\href{https://zenodo.org/record/4498861\#.YXU2A3X7R-M}{AI4D-Baamtu}, \href{https://huggingface.co/datasets/galsenai/WaxalNLP}{WaxalNLP}}, \textcolor{teal}{\href{https://zenodo.org/record/7561858}{Keyword Spotting}} & \href{https://github.com/samuelbenga/wolof-keyboard}{Wolof keyboard}, \href{https://github.com/srheal/Wol_Keyboards}{Wol\_keyboards}, \textcolor{green}{\href{https://github.com/abdouaziz/wolof}{Wolof library}}, \textcolor{cyan}{\href{https://stanfordnlp.github.io/stanza/}{Stanza}, \href{https://github.com/catherinearnett/morphscore}{MorphScore}}, \textcolor{brown}{\href{https://commonvoice.mozilla.org/wo}{Common Voice}, \href{https://www.dvoice.africa/}{Dvoice}}, \textcolor{magenta}{\href{https://github.com/UBC-NLP/AfroScope}{AfroScope}, \href{https://github.com/UBC-NLP/afrolid}{AfroLID}, \href{https://huggingface.co/PleIAs/CommonLingua}{CommonLingua},\href{https://github.com/cisnlp/GlotLID}{GlotLID}} \\ \midrule
Pulaar & \href{https://huggingface.co/datasets/TutlaytAI/AjamiXTranslit}{AjamiXTranslit}, \textcolor{purple}{\href{https://huggingface.co/datasets/HuggingFaceFW/fineweb-2/viewer/fuf_Latn}{FineWeb2}}, \textcolor{red}{\href{https://www.kaggle.com/datasets/michaelanietie/fulani-english-pair-dataset}{Fulani-English Pair Dataset}, \href{https://github.com/google-research/google-research/tree/master/madlad_400}{MADLAD-400}, \href{https://huggingface.co/datasets/google/smol}{SMOL}, \href{https://huggingface.co/datasets/HuggingFaceFW/finetranslations}{FineTranslations}} &\textcolor{brown}{\href{https://github.com/gauthelo/kallaama-speech-dataset}{Kallaama}, \href{https://huggingface.co/datasets/google/WaxalNLP}{WaxalNLP}}, \textcolor{teal}{\href{https://zenodo.org/record/7561858}{Keyword Spotting}}, \textcolor{violet}{ \href{https://huggingface.co/datasets/google/WaxalNLP}{WaxalNLP}} & {\centering -\par} \\ \midrule
Sereer & {\centering -\par} &\textcolor{brown}{\href{https://github.com/gauthelo/kallaama-speech-dataset}{Kallaama}}, \textcolor{teal}{\href{https://zenodo.org/record/7561858}{Keyword Spotting}} & {\centering -\par} \\ \midrule
Joola & {\centering -\par} &\textcolor{teal}{\href{https://zenodo.org/record/7561858}{Keyword Spotting}}& {\centering -\par} \\ \midrule
Mandinka & {\centering -\par} &\textcolor{teal}{\href{https://zenodo.org/record/7561858}{Keyword Spotting}}& {\centering -\par} \\ \midrule
Soninké & \href{https://huggingface.co/datasets/TutlaytAI/AjamiXTranslit}{AjamiXTranslit} & \textcolor{teal}{\href{https://zenodo.org/record/7561858}{Keyword Spotting}}& {\centering -\par} \\
\bottomrule
\end{tabularx}
\end{table}

\section{Current NLP Efforts and Tasks}\label{sec:effort}
\begin{figure*}[!htbp]
\centering
\includegraphics[width=1\textwidth]{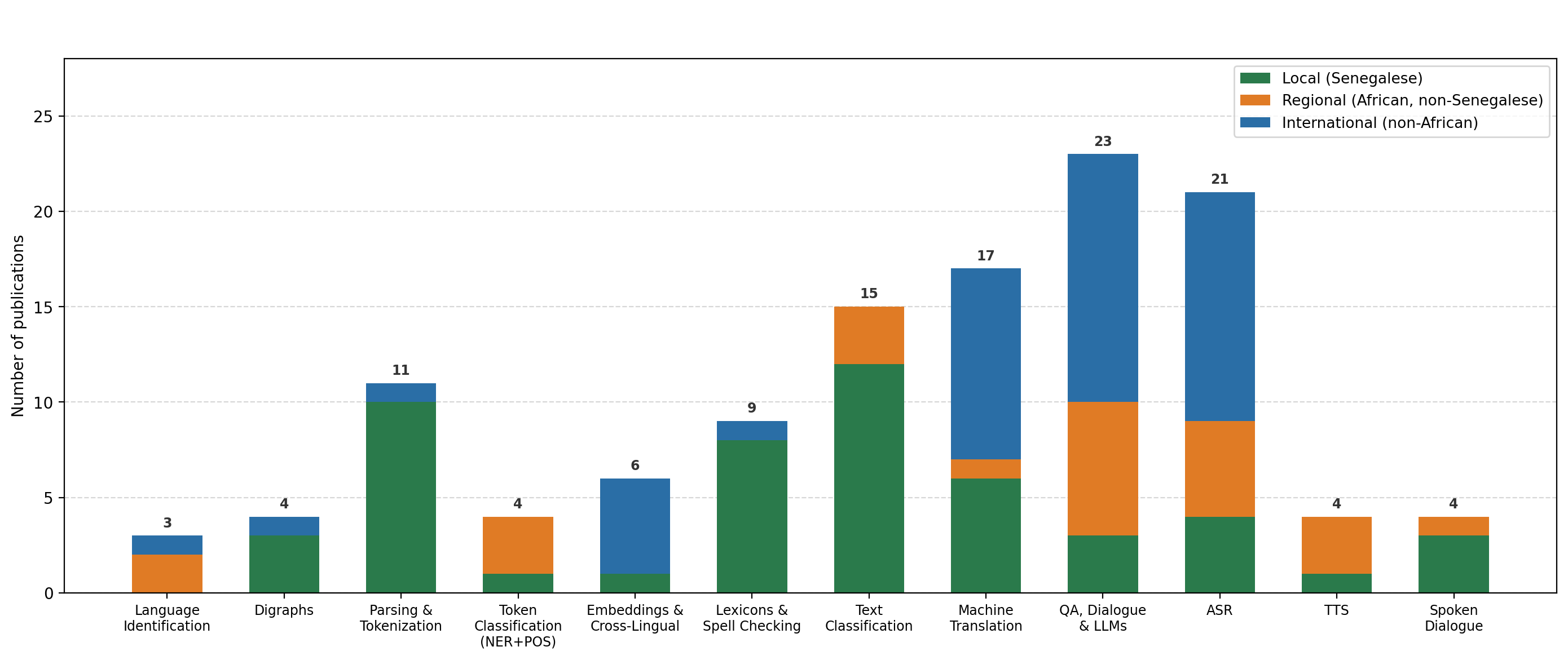}
\caption{Number of publications per NLP task category, broken down by research origin (Local, Regional, International). }
\label{fig:task_origins}
\end{figure*}

Senegal’s participation in African and global NLP initiatives is steadily growing, with empirical studies and open datasets focusing on the country’s main national languages. Figure~\ref{fig:task_origins} presents the collected publications organized by the various NLP tasks, depending on whether they were produced \texttt{locally (Senegalese)}, \texttt{regionally (African-non-Senegalese)}, or \texttt{internationally (non-African)}. The  Local researchers dominate \texttt{language-specific tasks} requiring community knowledge (parsing, lexicons, spell checking, text classification), while \texttt{international teams} lead infrastructure tasks included in large multilingual pipelines (ASR, MT, QA and LLMs). 
Tasks with zero local papers (Language Identification, NER, Fact Checking, Pre-training) represent direct priority targets for the Senegalese NLP community. This breakdown reflects the challenges faced by local researchers, particularly in terms of research funding, access to infrastructure, and highly qualified talent, as discussed in section~\ref{sec:challenges}.

This section synthesizes the current state of progress across major NLP tasks that form the core building blocks for future applied research, particularly in multilingual and interdisciplinary contexts.

\subsection{Language Identification}

Language Identification (LID) is the task of automatically determining the language in which a piece of text is written. It serves as a critical preprocessing step in NLP pipelines, particularly for mining web-crawled data, curating multilingual corpora, and enabling downstream tasks such as tokenization, machine translation, and sentiment analysis \citep{adebara2022afrolid, Kargaran2023Glotlid, commonlingua}. For African languages, LID presents unique challenges, including limited training data, closely related language varieties, morphological complexity, code-switching, and noisy web data \citep{kwon2026afroscope}.

Addressing the scarcity of LID tools for African languages, \citep{adebara2022afrolid} introduced \texttt{AfroLID}, a neural LID toolkit supporting 517 African languages and language varieties across 14 language families and five orthographic systems. AfroLID is trained on a manually curated, multidomain web dataset spanning all 50 African countries and achieves an F1 score of 95.89 on blind test data. The authors further demonstrate its utility on Twitter data, comparing it against various competing tools, on which AfroLID consistently outperforms all of them. Building on the need for broader language coverage and multilingual reliability, \citep{Kargaran2023Glotlid} presented \texttt{GlotLID-M}, an open-source LID model covering 1665 languages, the majority of which are low-resource. GlotLID-M is trained on GlotLID-C, a carefully curated corpus of 289 million sentences spanning 1832 languages, sourced from domains such as Wikipedia, religious texts, and news sites. In comparative evaluations against CLD3, FT176, OpenLID, and NLLB, GlotLID-M achieves superior performance when balancing F1 score and false positive rate, particularly on the UDHR benchmark where it improves over competing models by more than 12\% absolute F1. The authors further analyze challenges specific to low-resource LID, including incorrect metadata, leakage from high-resource languages, and noisy data. Most recently, \citep{kwon2026afroscope} introduced \texttt{AfroScope}, a unified framework for African LID comprising \texttt{AfroScope-DATA}, a dataset covering 640 African languages across eight language families, seven scripts, and eight domains, and \texttt{AfroScope-MODELS}, a suite of LID models trained on this data. A hierarchical disambiguation approach using \texttt{AfroScope-MIRROR}, a contrastive embedding model, is proposed for resolving confusions among closely related languages and varieties, achieving a 1.57 macro-F1 improvement on the confusable subset. The framework also provides a systematic analysis of cross-lingual transfer effects and domain robustness, offering practical guidance for building and evaluating African LID systems. In a complementary direction focused on efficiency, \citep{commonlingua} released \texttt{CommonLingua}, a 2.35-million-parameter byte-level LID model covering 334 languages, developed by PleIAs in partnership with the GSMA's ``AI Language Models in Africa, by Africa, for Africa'' initiative \footnote{All of the activities within this project have been subsequently transferred to the \href{https://www.atlasumoja.ai}{ATLAS Umoja AI website}.}. CommonLingua operates directly on raw UTF-8 bytes without a tokenizer, making it inherently script-agnostic, and combines three causal Conv1D layers with a bidirectional attention layer using RoPE. Trained on Structured Wikipedia and Common Corpus, it gives particular attention to the long tail of low-resource languages, supporting 61 African languages including those with minimal digital coverage. Despite its extremely small footprint, CommonLingua achieves state-of-the-art results on the CommonLID benchmark, outperforming larger systems such as \texttt{OpenLID v2} and \texttt{fastText-218}. A critical assessment of LID reliability has also been conducted in \citep{chen-etal-2024-fumbling}, where the authors systematically evaluated \texttt{ChatGPT}'s language identification capabilities across 670 languages. Their analysis reveals that while ChatGPT performs well on high-resource languages, it exhibits severe degradation on low-resource and African languages; frequently misidentifying, hallucinating, or refusing to respond; with performance on languages like Wolof falling well below acceptable thresholds. These findings underscore that LID remains an open challenge even for state-of-the-art LLMs, reinforcing the importance of dedicated, open-source LID tools such as those reviewed above for robust multilingual pipelines. Taken together, the LID literature shows a clear trajectory from feature-based systems to massively multilingual neural models, yet a reliable identifier for Senegalese national languages beyond Wolof remains absent; making language identification a bottleneck that propagates into every downstream task reviewed in this section.

\subsection{Parsing \& Tokenization}
Parsing is a foundational component of modern NLP, enabling systems to read, generate, and interpret text with unprecedented accuracy. It consists of analyzing the grammatical structure of a \texttt{sentence} or \texttt{text}, and identifies the roles of words such as \texttt{nouns}, \texttt{verbs}, and \texttt{adjectives}, and maps the relationships that link them together \citep{parsing}. Parsing generally assumes that the input has already been \texttt{tokenized} i.e. broken down into tokens (words, subwords, or symbols). Traditional syntactic parsing (dependencies, constituency) for examples, relies on \texttt{Word-level tokenization} to know the structured units to attach syntactic roles to \citep{dep-parsing}. 

The design and implementation of a morphological analyzer for Wolof has been reported in \citep{dione-2012-morphological} in order to obtain a linguistically motivated tool using \texttt{finite-state} techniques. As a foundational step toward an LFG-based\footnote{Lexical Functional Grammar.} computational grammar for Wolof, the authors introduced a newly constructed \texttt{Finite-State Transducer} (FST) for the language and presented experimental evaluations assessing the analyzer’s performance across several statistical metrics. 
To cope with the challenging \texttt{clitics}\footnote{A morpheme that has syntactic characteristics of a word, but depends phonologically on another word or phrase.} phenomenon in Wolof, \citep{bambaDione2013} adopted a non-transformational approach grounded in LFG to avoid verb-movement rules and preserves lexical integrity. The study includes an implementation showing how LFG effectively captures the behavior of Wolof clitics in order to demonstrate its practicality. Therefore, a finite-state transducer (FST) designed to tokenize and normalize natural-text input for a large-scale Wolof LFG grammar has been presented in \citep{dione-2017}. An initial language-independent tokenizer proved insufficient as issues with multiword expressions, clitics, and normalization emerged. Integrating FST components resolved these problems, enabling the grammar to handle free text more effectively and improving overall parsing performance. \citet{dione-2020} described an LFG-based parsing system for Wolof that handles major grammatical constructions. The system relies on FST for tokenization and morphological analysis, supported by lexicons and robust parsing strategies such as fragmenting and skimming. The system demonstrated solid parsing coverage on real Wolof data and achieved competitive accuracy after manual disambiguation, with strong precision and an overall balanced performance. \citet{sulger-etal-2013} presented the development of a multilingual parallel \texttt{treebank} spanning 10 languages including Wolof from 06 language families. Built using deep LFG grammars created within the \texttt{ParGram} project, the system produces highly parallel syntactic analyses across languages, that form the foundation of a richly annotated treebank. The analyses capture a wide range of linguistic phenomena and all produced resources are publicly accessible. To accelerate the LFG parsing process, \citet{dione-2014-pruning} introduced an efficient method that includes a Constraint Grammar (CG) parser into a probabilistic context-free grammar. Experiments demonstrated substantial gains in efficiency and robustness when annotating Wolof running text. Authors then presented a set of techniques for handling ambiguity in LFG parsing of Wolof in \citep{dione-disambiguation} by addressing both theoretical and practical concerns. The study explored multiple avenues to build a large-scale Wolof grammar and developed strategies that enhance the grammar’s efficiency, robustness, and coverage. The first \texttt{Universal Dependencies} (UD) resource within the Northern Atlantic branch of the Niger-Congo family has been presented in \citep{dione-2019-developing}. Various challenges related to word segmentation for tokenization and the mapping of Part-of-speech (PoS) tags (cf Section \ref{tok-class}), morphological features, and dependency relations to existing Wolof annotation conventions are reported. Several characteristic constructions are also outlined as a basis for discussing broader UD guidelines. This work has had a huge impact, enabling Wolof to be supported in NLP tools such as \texttt{Stanza} \citep{stanza}, or in larger studies such as \citep{arnett2025}, which evaluated Morphological Alignment of Tokenizers in 70 Languages. Authors introduced a tokenizer evaluation metric named \texttt{Morphscore}, which assesses tokenizers morphological alignment and help fill the performance gap between agglutinative languages like Wolof and fusional languages like English. A systematic method for deriving Universal Dependencies from LFG structures has been presented \citep{dione-2020-lfg}. Several challenges encountered by existing algorithms when applied to Wolof are discussed, along with the strategies adopted to address them. Evaluation results indicated that the approach achieved high accuracy and represented a clear improvement over earlier conversion methods. \citet{dione-2021-multilingual} leveraged multilingual embeddings and UD treebanks from both high-resource and low-resource languages including Wolof, to introduce a syntactic knowledge transfer method which allows to predict a wide range of UD annotations and dependency trees. Experiments showed that combining high-resource languages with low-resource languages in contemporary contact, leads to better performance than pairing unrelated languages. Overall, Wolof parsing research has followed a rule-based-to-neural arc: early FST and LFG grammars established the formal linguistic foundation, while the UD treebank \citep{dione-2019-developing} enabled integration into neural tools such as Stanza and MorphScore. This UD resource is the direct prerequisite for the token-level annotation tasks described in the next section, and the subword segmentation choices made during tokenization continue to propagate accuracy constraints into every downstream model.

\subsection{Token Classification}\label{tok-class}
Token classification is a natural language understanding (NLU) task where a specific label is assigned to each token (word or sub-word) in a text. It is used for tasks such as \texttt{Named Entity Recognition (NER)} to identify names, dates, and places, and \texttt{Part-of-Speech (POS)} tagging to categorize words as nouns, verbs, adjectives, and so on \citep{token-classification}. 

Regarding NER, \citep{adelani-etal-2021-masakhaner} introduced the first large, publicly available, high-quality dataset in 10 African languages including Wolof\footnote{The coverage has subsequently been extended to 20 African languages in \citep{adelani-etal-2022-masakhaner} with no additional Senegalese languages.}. Authors train and evaluate multiple NER models and conducted an extensive empirical evaluation of state-of-the-art methods across both supervised and transfer learning settings. A complementary model-side approach to NER in languages absent from standard pretrained models has been proposed in \citep{e-mbert}, where the authors introduced \texttt{E-MBERT}, a method to extend Multilingual BERT (M-BERT) to new languages via vocabulary enlargement and continued pre-training on monolingual data. Evaluated across 27 languages, including 11 not covered by M-BERT (such as Wolof, Hausa, and Akan), \texttt{E-MBERT} achieves an average F1 increase of 23\% on new languages and 6\% on languages already in M-BERT, offering a practical and efficient solution for languages such as those spoken in Senegal that remain outside standard multilingual model coverage. From a comparative standpoint, the two strategies represent complementary trade-offs: cross-lingual transfer from MasakhaNER leverages a purpose-built African-language annotation scheme and provides Wolof-specific evaluation data, while E-MBERT's vocabulary extension achieves broader language coverage without requiring any Wolof-annotated NER data at training time. Neither approach, however, has been directly benchmarked against the other on the same Wolof test split, and the relatively small size of the MasakhaNER Wolof test set means that reported F1 scores carry non-trivial variance. Future work establishing a common held-out benchmark would be essential to properly rank these approaches and to determine whether the performance gap observed for Wolof NER stems from architecture, vocabulary coverage, or annotation quality. Similarly, very little work has been done in POS tagging in Senegalese languages. The design of a part-of-speech-tagset for Wolof has been reported in \citep{dione-etal-2010-design} alongside with an efficient process of creating a semi-automatically annotated gold standard. Authors leveraged available lexical resources and used purpose-built heuristic tools for stemming and guessing of word forms. They evaluated afterwards the performance of state-of-the-art statistical PoS taggers on the collected data, and briefly summarize cross-lingual projection experiments utilizing the parallel corpus data. \texttt{AfricaPOS}, the largest part-of-speech (POS) dataset for 20 typologically diverse African languages including Wolof has been introduced in \citep{dione-etal-2023-masakhapos}. Researchers conducted extensive POS baseline experiments using both Conditional Random Field (CRF) and several multilingual pre-trained language models and discussed the challenges in annotating POS for these languages using the universal dependencies (UD) guidelines. All resources (data, code, and models) have been released to inspire future research in NLP on these languages. Collectively, these efforts establish Wolof as the only Senegalese language with published NER and POS resources, while Pulaar, Sérère, Diola, Mandingue, and Soninké remain entirely uncovered at the token classification level. This asymmetry has direct consequences for the text classification models discussed next: cross-lingual transfer from Wolof to other national languages cannot be properly evaluated without equivalent annotated benchmarks.

\subsection{Text Classification}
Unlike token classification which assigns a label to each token, text classification assigns a single label to an entire sentence. It is a core task in NLP used for applications like \texttt{spam filtering}, \texttt{sentiment analysis}, and \texttt{news categorization} to organize and analyze large volumes of unstructured text \citep{text-classification}.

\subsubsection{Sentiment Analysis}\label{sec:sentiment}
Sentiment analysis (also known as opinion mining) is the process of using NLP to identify and extract subjective information from text to determine the author's emotional tone as \texttt{positive}, \texttt{negative}, or \texttt{neutral} \citep{sentiment-analysis}. It is very helpful to understand public opinion on products, services, and brands by analyzing data like customer reviews, support tickets, and social media comments. \citet{term-weighting} conducted a survey on the term weighting schemes as it represents the crucial step for representing the documents in a suitable way for classification algorithms. Authors proposed an efficient term weighting scheme that provide better classification accuracy than \texttt{TF-IDF} \citep{tfidffeature} and \texttt{IF-IGM} \citep{tfigm}. As code-switching is quite common in Senegalese languages with the presence of French \citep{beqi}, an extended lexicon with French and Wolof words and expressions used in both languages was proposed in \citep{fwlsa-score}. Researchers introduced a sentiment scoring algorithm named \texttt{FWLSA-score}\footnote{FWLSA = French and Wolof Lexicon-based Sentiment Analysis.}, that uses word similarity to address the spelling problem, and classifies reviews as positive or negative based on the polarity of the words or expressions. An improvement over the FWLSA-score has subsequently been studied in \citep{trigram} with word-level trigrams (list of consecutive three letters) in order to improve its effectiveness on verbs conjugation and words declination in both languages. However, relying on word-level similarity and trigrams to map inflected/derived forms to a base form, are complex and limited, especially for morphologically complex Wolof and French words written using the French alphabet. To advance bilingual French-Wolof sentiment analysis, the authors in \citep{markov} proposed \texttt{(i)} a new French-Wolof dictionary and dictionary-based lemmatizer to accurately handle morphology, and \texttt{(ii)} a novel \texttt{Markov Model-based} method for identifying context-dependent sentiment words. To address the complexity related to ambiguity in Senegalese on-line press comments, \citep{senopinion} suggested an opinion lexicon with Wolof, French and urban language words and expressions, to process these types of data. Authors therefore aim to pave the way for the development of tools for processing local languages. Another approach, based on \texttt{graph structures} has been studied in \citep{graph} to address the same challenge. Researchers proposed a dictionary-based lemmatizer and an graph algorithm to model the relationship between French and Wolof opinion words.
\texttt{COMFO}, a multilingual corpus (French, English, and Urban Wolof) for opinion mining, has been introduced in \citep{faty-etal-2022-comfo} to facilitate the exploration of supervised learning algorithms for sentiment classification. The authors detailed the corpus collection process, covering the data source, preparation, and the lexical-based annotation approach used. In \citep{young-sentiment} we collected a set of \texttt{X} (formely  \texttt{Twitter}) and \texttt{Facebook} comments related to the youth’s perception about the mobile internet costs in Senegal and applied sentiment analysis to gather their general feelings. We leveraged a multilingual language model based on XLM-Roberta \citep{conneau-etal-2020} and pre-trained on nearly 200 million Tweets across some 30 languages (including French). Domain-specific model (in this case, social media) is more effective than its general counterpart when it comes to refining task-specific multilingual Language Models \citep{barbieri2022}. The progression from lexicon-based (FWLSA-score, trigrams, Markov models) to neural (XLM-RoBERTa) approaches reflects a clear trajectory, but no study has systematically benchmarked these systems on the same held-out test set with reported confidence intervals, making it impossible to quantify how much of the apparent gain comes from model expressiveness versus corpus size or domain coverage. Most datasets are drawn from a single social media platform (Twitter or Facebook), constraining generalizability to other registers such as the agricultural or healthcare discourse.

\subsubsection{Hate Speech Detection}
The immense growth and public nature of social media, where any content can be posted and reach millions, necessitates automated methods for identifying inappropriate content. Among these, the detection of \texttt{hate speech} is crucial, despite its complex and subjective definition \citep{malik2023deep}. Given the scale of social media, the systems used to detect hate speech must be highly accurate, effective, and efficient which is especially challenging in low-resource languages. Researchers in \citep{jacobs2023} investigated hate speech detection in low-resource languages through the lens of \texttt{KeyWord Spotting (KWS)} (cf Section \ref{sds}). The main objective is to search through an audio corpus for a pre-determined set of keywords indicative of hate speech. Their findings suggested that KWS using multilingual \texttt{Acoustic Word Embeddings (AWEs)} is a promising approach for quickly implementing hate speech detection in a new unseen language (here Wolof and Swahili) if resources are severely limited. To contribute to lowering this resource barrier, \citep{awofro} introduced \texttt{AWOFRO}: the first open annotated corpus of 3510 tweets in code mixed Wolof. Authors performed an exploratory analysis of the corpus and validated the annotations using \texttt{Cohen's Kappa} measures. A comparative study of machine learning models for abusive message detection has been presented in \citep{ndao:hal} focusing on code-mixed\footnote{The use of two or more languages in the same sentence.} data in Wolof and French languages. Authors introduced a meticulously annotated dataset of 2,022 tweets, which were manually classified as abusive or non-abusive. They also conducted extensive experiments, comparing the performance of various machine learning and deep learning algorithms on this dataset. \citet{abuse-bert} introduced \texttt{AbuseBERTWoFr}, the first model for abusive message detection on Wolof-French code-mixed tweets, trained on a large dataset of nearly 145k tweets. Researchers evaluated the model's performance on a corpus of +2k code-mixed tweets, and then compared its results against state-of-the-art language models. While these results are encouraging, the small evaluation corpus (2k samples) and the absence of cross-dataset validation mean that reported performance numbers may overestimate generalization: a model trained and evaluated on the same Twitter sampling period is susceptible to temporal and topical distribution shift when applied to new data streams or different social media platforms.

\subsubsection{Intent Classification}
In Task-Oriented Dialogue (ToD) systems (cf Section \ref{dialog}), Natural Language Understanding (NLU) is essential for identifying the user's \texttt{main goals} and \texttt{information} \citep{razumovskaia2022}. NLU is typically split into two sub-tasks: 

\begin{itemize}
    \item \texttt{Intent classification} which consists of assigning one or more goal labels to the entire utterance \citep{ravuri2015};
    \item \texttt{Slot filling} in which specific values are extracted from the utterance to populate predefined information slots \citep{mesnil2014}.
\end{itemize}

Although operating at different levels (token vs. sentence level), they are generally performed as a \texttt{joint task} to maximize performance in both simultaneously \citep{weld2021survey}. Similarly to other NLP tasks, existing large-scale benchmarks often omit low-resource languages and tend to heavily rely on English translations, which results in a predominant focus on Western-centric concepts. To mitigate this limit, \texttt{\textsc{Injongo}}, a multicultural open-source benchmark dataset for 16 African languages including Wolof, has been introduced in \citep{injongo}. Authors covered 05 domains\footnote{Banking, Home, Travel, Utility, and Kitchen \& Dining.} and performed several supervised fine-tuning experiments with multilingual encoders and Large Language Model (LLM) prompting. \citet{kandji2025wolbanking77} introduced an intent classification dataset consisting of around 10k customer service queries (Bank and Transport) from the \texttt{Banking77} dataset \citep{casanueva} translated to French and Wolof. Authors evaluated different pre-trained models in \texttt{zero-shot} and \texttt{few-shot} settings and reported promising results. However, these results may be biased due to labeling errors that were discovered in the original \texttt{Banking77} dataset \citep{ying-thomas-2022-label}. Furthermore, it has been observed that translationese\footnote{Texts translated by either humans or machines.} often exhibits features such as \texttt{stylistic ones} that are different from text written directly in the original language and thus can mislead model training \citep{yu-etal-2022-translate}. This is a major problem in African language datasets, which are mainly based on translations of existing corpora \citep{singh-etal-2024-aya}. To date, this is the only papers that specifically addresses intent classification in African languages, including at least one Senegalese language. This task is also studied in the works presented in Section \ref{dialog}, in the ToD context.

\subsubsection{Fact Checking}
Automated fact checking (AFC) is the task of verifying the truthfulness of a claim using computational methods, typically as a multi-stage pipeline involving \texttt{information retrieval} (finding relevant documents), \texttt{evidence extraction} (identifying sentence-level passages that support or refute the claim), and \texttt{claim verification} (classifying the claim as \textit{Supported}, \textit{Refuted}, or \textit{Not Enough Information}) \citep{azime2026}. This task is of particular importance in the social science context, as misinformation about health, culture, and public affairs disproportionately affects communities with limited access to information; precisely the situation of speakers of low-resource languages. \citet{azime2026} introduced \texttt{AfrIFact}, the first multilingual AFC benchmark explicitly targeting African languages, covering \textbf{10 languages including Wolof}, across health and culture-news domains. The dataset comprises more than 18,000 claims with annotated evidence spans, constructed through a rigorous native-speaker annotation pipeline. Evaluation of state-of-the-art embedding models and LLMs reveals three key findings: (i) even the best embedding models lack reliable cross-lingual retrieval capabilities for African languages; (ii) cultural and news documents are easier to retrieve than healthcare documents; and (iii) LLMs exhibit poor multilingual fact-verification performance, though few-shot prompting improves results by up to 43\% (AfriqueQwen-14B \citep{afriquellm}) and task-specific fine-tuning yields an additional 26\% gain. The dataset and code are publicly available, providing a foundation for future work on low-resource information retrieval, evidence retrieval, and fact checking; tasks with direct applications for combating misinformation in Senegalese language communities.

Across the four text classification sub-tasks reviewed here, a consistent pattern emerges: lexicon-based and classical ML approaches were pioneered for Wolof-French code-switched data, while more recent work increasingly relies on cross-lingual transfer from multilingual pretrained models, with French generally serving as a more effective pivot than English for these languages. Datasets remain Wolof-centric and small-scale, and the absence of any annotated text classification corpus for Pulaar, Sérère, or the remaining three national languages means that the community has not yet tested whether these findings generalize. The shared dependence on social media data (Twitter, Facebook) also introduces domain and demographic biases that should be addressed in future benchmark design.

\subsection{Lexicons \& Spell Checking}
Writing clearly and accurately can be challenging, especially for non-native speakers, as there are often many ways to express the same idea. A single spelling error (unexpected word form) significantly hinders readability and processing. In applications like Natural Language Processing (NLP), unnormalized, incorrectly spelled, or poorly digitized text severely diminishes its informational value \citep{electronics9101670}. To overcome the writer's constraints of time and proficiency, Automatic Spelling Correction (ASC) is deployed to locate misspelled words and generate a ranked set of potential replacements. Several approaches have therefore been studied to solve the problem of automatic spell checking. The study conducted in \citep{survey-spell-check} divides these approaches into three categories:

\begin{itemize}
    \item Those based on \texttt{expert rules};
    
    \item Those incorporating a \texttt{context model} that allows candidate corrections to be reorganized;
    
    \item Those that learn \texttt{error patterns from a training dataset}.
\end{itemize}

Although significant progress has been made in the field of spell checking for low-resource languages, little work has been done specifically for Wolof. As part of the African Language-French Dictionaries (DiLAF) project, several dictionaries covering five other African languages in addition to Wolof have been developed \citep{mangeot:hal-01107550} whose online publication was presented in \citep{Cheikh2015}. The implementation of a spell checker for Wolof has been studied in \citep{lo:hal-02054917}, using a lexical approach based on a French-Wolof dictionary \citep{Khoule2016iBaatukaay} and a Wolof morphological analyzer \citep{dione-2012-morphological}. However, this work did not go as far as implementing a functional spell checker and was limited to a review of existing methods based on expert rules and context models based on n-gram language models. Furthermore, at the time of writing, all the dictionaries developed in \citep{Cheikh2015} are available online, apart from the Wolof dictionary\footnote{\url{http://pagesperso.ls2n.fr/~enguehard-c/DiLAF/index.php}}. This absence prevents the exploration of dictionary-based approaches, even though these latter present several limitations:

\begin{itemize}
    \item The \texttt{maintenance complexity} due to the rapid increase in the number of rules and the increasing difficulty of updates ;

    \item The dependence on the size of the dictionary ;
    
    \item The lack of linguistic context awareness.
\end{itemize}

A proper Wolof spell checker has been proposed in \citep{cisse-sadat-2023-automatic} and relies on a combination of trie data structures, dynamic programming, and weighted Levenshtein distance to generate suggestions for misspelled words. The authors created new linguistic resources for Wolof i.e., a lexicon and a corpus of misspelled words, using a semi-automatic approach that combines manual and automatic annotation methods. However, the correction techniques described therein focus exclusively on the word level and do not take into account the context in which it appears. This limitation is especially acute for Wolof's agglutinative morphology: because multiple grammatical morphemes can be concatenated into a single orthographic word, an incorrectly formed suffix may produce a string that passes a dictionary lookup or resembles a legitimate word form, causing a word-level corrector to either accept an erroneous form silently or propose a morphologically implausible candidate. In inflectionally simpler languages such as French or English, the same failure mode occurs less frequently because the inflectional paradigm is smaller and suffix ambiguity is lower. This means that the effective error coverage of a Wolof word-level corrector is systematically narrower than headline metrics suggest, and that context-aware or morphology-aware correction is a higher priority for Wolof than for most European-language baselines. The integration of a context model, usually an n-gram language model \citep{langmodel}, nevertheless allows contextual information to be included based on the history of previous words. However, this approach remains limited, as it only takes into account the immediate context preceding a word. Although additional classifiers can be used to overcome these limitations \citep{survey-spell-check}, the use of neural networks allows for the integration of a broader context, taking into account the words on both sides of the target word. Thus, deep learning with neural networks with attention \citep{bahdanau} is a promising approach, which has already been studied for spell checking in various languages. This approach addresses this task by modeling spelling correction as a \texttt{translation task} from misspelled (noisy) text to well-spelled (correct) text and shows promising results. However, it requires a parallel corpus of noisy data on the one hand and correct data on the other hand, whereas languages like Wolof are low-resource languages and might not have such a corpus. In \citep{beqi}, we introduced \textsc{Beqi}: an efficient way to address this constraint by generating synthetic data based on regex\footnote{A string of characters that defines a search pattern for matching text} and seed data scraped on social media. We presented sequence-to-sequence models based on LSTM and Transformers for spelling correction in Wolof and evaluated these models in three different scenarios depending on the subwording method applied to the data. The work in \citep{cisse-sadat-2024-advancing} followed the same direction by leveraging transformer models and neural networks for word correction and spelling in Wolof. Authors also introduced a model trained on a parallel corpus consisting of misspelled sentences and their error-free counterparts and optimized the model to translate error-prone text into accurate sentences. Although the datasets used in \citep{beqi} and in \citep{cisse-sadat-2024-advancing} had not been published at the time of writing, \citet{oolel-corrector-data} released a dataset of \texttt{3,400 sentence pairs}: informal Wolof on one side, standard Wolof on the other\footnote{Publicly available at \url{https://huggingface.co/datasets/soynade-research/Wolof-Non-Standard-Orthography}.} with the aim of helping models understand Wolof as it's actually written online. They subsequently fine-tuned a smaller version of \citep{oolel} trained to convert informal, social-media Wolof into standard orthography, while leaving code-switched French and English segments untouched \citep{oolel-corrector}.

Text normalization is a foundational step in text processing under-resourced languages, especially those with inconsistent orthographies. A centralized and up-to-date lexical database is therefore essential for defining a common reference system. The iBaatukaay project \citep{khoule:hal-01992863} has been initiated as a collaborative project whose objective was to design a collaborative multilingual lexical database on the web for African languages, particularly Senegalese ones. Any expert in the field (lexicographers, linguists, etc.) could contribute via Internet, and the data could be downloaded free of charge from the platform. The project presented 25 indigenous senegalese languages, three of which were chosen for the project's launch: Wolof, Pulaar, and Bambara. Nevertheless, the \texttt{Institut Fondamental d'Afrique Noire (IFAN)} and the \texttt{École Supérieure Polytechnique (ESP)} went further and launched \textsc{Sentermino} \citep{sentermino} : a terminology platform aimed at centralizing, harmonizing, and providing scientific and technical terminology that has been validated and adapted to national languages. Such a platform facilitates the production of scientific and educational content in national languages, which improves the availability of data in local languages online. It also harmonizes the use of terms referring to the same concepts, which mitigates code-switching \citep{cetinoglu-etal-2016-challenges} and reduces the vocabulary of NLP systems to maximize their performance \citep{outeirinho-etal-2024-exploring}. Taken together, the spell-checking and lexicon efforts reviewed here address the same core bottleneck that constrains machine translation (Section~\ref{sec:mt}) and ASR language modelling (Section~\ref{sec:asr}): the absence of a clean, standardized written reference for Wolof. Robust spell correction is not only a standalone tool but a critical preprocessing stage for MT and ASR pipelines; noisy orthographic input degrades translation quality and increases word error rate, making the two research agendas mutually reinforcing.

\subsection{Machine Translation}\label{sec:mt}
A machine translation (MT) system converts a text sequence (or audio source) from a source language into the same sequence in a target language. For a long time, statistical machine translation (SMT) systems \citep{koehn_2009} were the dominant approach before the emergence of neural machine translation (NMT) systems \citep{luong-etal-2015-effective}, which have gradually achieved increasingly higher performance.

However, the quality of these systems has always been closely linked to the amount of data used in their design \citep{koehn-knowles-2017-six}. Therefore, \texttt{NTREX-128}, a dataset for machine translation evaluation from English into a total of 128 target languages has been released in \citep{federmann-etal-2022-ntrex}, comprising around 2k sentences for each language including Wolof. \citet{caswell2025} open-sourced \texttt{SMOL (Set of Maximal Overall Leverage)}, a suite of 6.1M tokens training data that has been translated into 124 (and growing) under-resourced languages (125 language pairs including Wolof and Pulaar), including many for which there exist no previous public resources. These initiatives are very important as they contribute to unlock machine translation for low-resource languages. Thus, the most advanced machine translation systems have been developed with sequence-to-sequence models exploiting the attention mechanism \citep{bahdanau} as well as the Transformer architecture \citep{NIPS2017_3f5ee243}. Neural machine translation for low-resource languages (LRL-NMT) has been the subject of extensive research within the community, and various approaches have been studied. An overview of LRL-NMT work has been provided in \citep{ranathunga2021neural}, along with a set of recommendations for optimizing the design of translation systems based on the configuration of the language data (size, type of datasets, and available computational resources). Despite the substantial work carried out in neural machine translation in low-resource languages, very few local studies have specifically targeted Senegalese languages. To our knowledge, the first studies that have specifically explored Wolof-French machine translation systems are those presented in \citep{nguer-etal-2020-sencorpus}, where the authors introduced a corpus of 70,000 Wolof-French parallel sentences used to develop Word Embedding models \citep{alla-word-embeddings} as well as translation models \citep{alla2020usinglstmtranslatefrench} based on the LSTM architecture \citep{lstm}. 
However, the results presented in \citep{nguer-etal-2020-sencorpus} were evaluated in terms of Accuracy, which complicates the effective assessment of the translation quality of their systems. We addressed this gap in \citep{derguenelstm}, where we proposed an LSTM-based machine translation system that is evaluated using the BLEU metric \citep{papineni-etal-2002-bleu}. BLEU is commonly used in the evaluation of NMT systems and offers a better correlation with human evaluations than Accuracy \citep{medium-nmt}. \citet{alla2020usinglstmtranslatefrench} also used the BLEU metric, but their results were biased due to a significant overlap between the training, validation, and test sets; an issue known as \textsc{data leakage}. \citet{dione-etal-2022-low} indicate that approximately 60\% of the sentences in the test set were also found in the training data, which greatly overestimated the model's capabilities. Data leakage is one of the pitfalls associated with the adoption of machine learning methods, leading to failures in terms of validity, reproducibility, and generalization \citep{kapoor2023reformsreportingstandardsmachine}. \citet{lones2024avoidmachinelearningpitfalls} provides a set of best practices for avoiding these errors, ranging from what to do before building the model, to how to build reliable models, evaluate them robustly, compare models fairly, and report results. The corpus initially introduced in \citep{nguer-etal-2020-sencorpus} has thereafter been subsequently expanded to 83,000 sentences in \citep{dione-etal-2022-low}, enabling the training of two neural machine translation systems for the French$\rightarrow$Wolof and Wolof$\rightarrow$French directions, based on the \textsc{Transformer} architecture \citep{NIPS2017_3f5ee243}. A translation platform named \texttt{\textsc{SenTekki}} \citep{sentekki} has subsequently been set up based on this model, to allow the public to interact with the system via a web interface when deployed. A RestFul Web Service\footnote{A service that uses the principles of REST (Representational State Transfer) to allow applications to communicate over the web using standard HTTP methods. } was also developed, enabling other applications to integrate translation features, which is very important for the effective inclusion of our local languages. However, the authors did not mention the deployment of the platform, and no URL was provided to access it. It is also interesting to note that none of the French-Wolof datasets mentioned above have been made publicly available to date. This makes reproducibility difficult and hinders the progress of local work on these languages. This lack of openness could be explained by the still low level of research funding in Senegal and more broadly across the African continent (less than 1\% of GDP). Comparing the local MT results that report clean BLEU evaluations reveals that performance is primarily corpus-size-driven rather than architecture-driven: the best published result for French$\rightarrow$Wolof on a cleanly split evaluation is \textbf{26.38 BLEU} \citep{mbaye-diallo-2025-task}, obtained by fine-tuning SMALL-100 on 175k parallel sentences. By contrast, earlier Transformer models trained on 83k sentences \citep{dione-etal-2022-low} achieved lower scores, and LSTM baselines on 70k sentences \citep{derguenelstm} were weaker still. This progression is fully explained by corpus size and pretrained backbone quality; the SMALL-100 model benefits from massively multilingual pretraining before Wolof fine-tuning, making it impossible to credit the gain to architecture alone. The remaining bottleneck is therefore not a modelling gap but a data gap: domain-specific parallel corpora in health, legal, and educational registers would yield larger practical gains than further scaling of general-domain training data.

Pre-trained multilingual translation models are also an interesting direction which have shown promising performance in supporting low-resource languages. They enable information sharing between similar languages, which significantly improves the translation of these language pairs, as studied in \citep{arivazhagan2019massively}. Much work has therefore been done in this direction, leading to the development of a wide range of multilingual translation models that include at least, the Wolof language. Such models have been developed in \citep{adelani-etal-2022-thousand}, where the authors leveraged existing pre-trained models to design translation systems for 16 low-resource African languages. Complementing these end-to-end approaches, \citep{bitext-mining} proposed \texttt{LASER2}/\texttt{LASER3}, a teacher-student framework that trains language-family-specific sentence encoders for 44 African languages, enabling large-scale bitext mining against 21.5 billion English sentences. The mined bitexts yielded a +3.8 BLEU improvement for neural machine translation into English, demonstrating that high-quality multilingual sentence representations are a powerful lever for bootstrapping parallel data in low-resource settings. Taking an Afrocentric direction, \citep{elmadany-etal-2024-toucan} first pretrained two sequence-to-sequence language models, \texttt{Cheetah-1.2B} and \texttt{Cheetah-3.7B}, on 517 African languages, then fine-tuned them to create \texttt{Toucan}, a many-to-many MT model supporting 156 language pairs across 43 African languages plus Arabic, English, and French. The authors also introduced \texttt{AfroLingu-MT}, the largest African MT benchmark to date, and \texttt{spBLEU1K}, a sentencepiece-based evaluation metric covering 1,003 languages including 614 African ones, providing both stronger models and more reliable evaluation tools for the community. Meta (formerly Facebook) introduced \textsc{M2M-100} \citep{fan2020englishcentric} as the first multilingual machine translation model capable of translating between any pair of 100 languages without relying on English data. A distilled version of M2M100 named \textsc{SMALL-100} was introduced in \citep{mohammadshahi2022small100} with the particularity of being \texttt{3.6 times smaller and 4.3 times faster} at inference while having equivalent performance. The Wolof$\leftrightarrow$French dataset introduced in \citep{derguenelstm} has been expanded to \textbf{175,000 sentences} and then used in \citep{mbaye-diallo-2025-task} to fine-tune the SMALL-100 model, achieving a BLEU score of \textbf{\texttt{26.38}}. This is to date the largest locally created French$\leftrightarrow$Wolof corpus (not openly available too). In a complementary effort focused on efficient deployment, \citep{moslem-etal-2026-afrinllb} introduced \texttt{AfriNLLB}, a series of lightweight MT models obtained by compressing \texttt{NLLB-200 600M} via iterative layer pruning and quantization, covering 15 language pairs including both English$\leftrightarrow$Wolof and French$\leftrightarrow$Wolof. Fine-tuned with knowledge distillation from a larger teacher model on curated parallel corpora, \texttt{AfriNLLB} achieves performance comparable to the full baseline while being significantly faster, enabling deployment in resource-constrained settings. Both the models and all training data are publicly released. The work on M2M100 has also subsequently been expanded by META upon in the \textsc{No Language Left Behind} project\footnote{\url{https://ai.facebook.com/research/no-language-left-behind/}} offering a state-of-the-art model capable of translating 200 languages into each other \citep{nllbteam2022}. Building on this trajectory, \citep{omnilingualsonarteam} introduced \texttt{OmniSONAR}, a unified cross-lingual and cross-modal sentence embedding space natively covering 4,200 language varieties. Trained via a progressive teacher-student distillation strategy, \texttt{OmniSONAR} achieves a 15-fold error rate reduction over prior best models across 1,560 languages in the BIBLE benchmark and outperforms NLLB-3B on several multilingual translation benchmarks, establishing a new state of the art in multilingual representation learning at scale. Building directly on OmniSONAR, \citep{omnilingualmtteam2026} introduced \texttt{Omnilingual MT (OMT)}, the first MT system supporting more than 1,600 languages, through two complementary architectures: \texttt{OMT-LLaMA}, a decoder-only model built on LLaMA3 with multilingual continual pretraining and retrieval-augmented translation, and \texttt{OMT-NLLB}, an encoder-decoder model that leverages OmniSONAR's aligned representations. All 1B to 8B parameter OMT models match or exceed the performance of a 70B LLM baseline, demonstrating a clear specialization advantage. OMT also introduces \texttt{BOUQuET}, a large manually created multilingual MT evaluation collection spanning a wide range of linguistic families, along with \texttt{BLASER 3} (a reference-free quality estimation model) and \texttt{OmniTOX} (a toxicity classifier), establishing a comprehensive evaluation framework for low-resource MT at scale. In 2024, the well-known translation platform \textsc{Google Translate}\footnote{\url{https://translate.google.com/?sl=fr&tl=wo&op=translate}} expanded its support for underrepresented languages thanks to its \textsc{Palm 2} large language model \citep{anil2023palm2technicalreport} to 110 additional languages, including Wolof \citep{google-translate-extension}. As this service is proprietary, they subsequently released \texttt{TranslateGemma} \citep{finkelstein2026}, a suite of open machine translation models based on the \texttt{Gemma 3} foundation models \citep{gemmateam2025}. Its multilingual translation capabilities are enhanced through a \textbf{two-stage fine-tuning process} combining \texttt{supervised learning} on large-scale synthetic and human-translated parallel data, followed by \texttt{reinforcement learning} optimized with multiple quality-focused reward models. The resulting TranslateGemma models demonstrate consistent and significant improvements over baseline Gemma 3 across a wide range of language pairs, including low-resource African languages such as Wolof and Fulani, while achieving competitive performance even at smaller model sizes. Strong multimodal translation capabilities are also retained, making the open release of TranslateGemma a versatile and efficient tool for the research community. The \texttt{\textsc{DeepL Translate}}\footnote{\url{https://www.deepl.com/en/translator}} platform, considered as the main alternative to Google Translate (although having less language support), now also offers Wolof \citep{deepllangsupport}. This is a major step forward as DeepL is often more accurate for nuanced translations and represents a better choice for professional use \citep{deepl-translate}. At the local level, players such as \texttt{Baamtu Technologies}\footnote{\url{https://baamtu.com/}}, a pioneering AI company in Senegal \citep{HENG2022102454}, and more recently \texttt{LAfricaMobile}\footnote{\url{https://lafricamobile.com/en/produit-tts/}}, were already offering proprietary machine translation systems in local languages. The academic world is also getting involved with the publication of \texttt{Firilma} \citep{firilma}, a French-Wolof translation system developed by the \texttt{Center for Applied Linguistics in Dakar (CLAD)}. The arrival of these new major stakeholders therefore represents both an opportunity and a threat, forcing local players to quickly reinvent themselves and adapt their strategic positions \citep{Osiris2024}. Three broad generations of MT work for Senegalese languages can be identified: locally trained sequence-to-sequence baselines on small parallel corpora (70k--175k sentences), multilingual fine-tuning of large pre-trained models (M2M100, NLLB, AfriNLLB), and proprietary tech-giant systems (Google Translate, DeepL, TranslateGemma). Performance improvements across these generations have been predominantly data-driven rather than architecture-driven, and the remaining bottleneck is the creation of domain-specific parallel corpora in health, legal, and educational registers rather than further model scaling. Crucially, \textbf{none of the locally created French--Wolof parallel datasets have been publicly released}, limiting reproducibility and constraining the community to rebuilding training data from scratch for each new effort. A notable exception to this broader pattern of closed resources is \texttt{MudawanSn} \citep{mbaye-diop2026gold}, the first locally created MT parallel corpus to be openly released. It is a gold-standard Wolof--Arabic dataset of 1,271 manually translated sentence pairs drawn from Senegalese news texts and verified through a LASER2 cross-lingual alignment loop (average cosine similarity of 0.75). Hosted on Hugging Face under a \texttt{CC BY-NC-SA 4.0} license, \texttt{MudawanSn} also opens a new translation direction Wolof$\leftrightarrow$Arabic, that is absent from prior local MT work, and demonstrates that producing and openly sharing a locally curated parallel corpus is a tractable goal even in a low-resource setting.

\subsection{Question Answering and Dialogue Systems}\label{dialog}
Dialogue systems allow users to interact in natural language, via \textbf{text} or \textbf{voice}, in order to perform specific tasks (e.g. make reservations, obtain information, order a service) or to engage in open conversation (chatbots). Modern practice frequently combines \texttt{task-oriented paradigms} (driven by frames/slots) and \texttt{open conversation} in the same assistant, such as Siri, Alexa, and Google Assistant, or more recently, with the rise of \textsc{Generative AI} and \textsc{LLMs}, products such as ChatGPT, Gemini, and Claude. Despite technological advances, dialogue modeling remains a major challenge. Conversational agents must be able to handle interactions on a wide range of topics, provide relevant responses, and adapt to varied linguistic and cultural contexts \citep{jurafsky}. In addition, dialogic phenomena (multi-turns, initiative, grounding, corrections) require a large volume of training data, robust and adaptive architectures \citep{young2013pomdp} and issues like ethics and algorithmic biases raise concerns about the accessibility and neutrality of models. LLMs are currently the de facto backbone of modern chatbots, and as with major advances in NLP and AI in general, low-resource languages remain underserved. \citet{adelani2024irokobench} introduced a human-translated benchmark dataset for 17 typologically diverse low-resource African languages, called \textsc{IrokoBench}. It covers three tasks: \texttt{Natural Language Inference}, \texttt{Mathematical Reasoning}, and \texttt{Multi-choice Knowledge-based QA}. Their evaluation of  10 open and 06 proprietary large language models (LLMs) in zero-shot, few-shot, and translate-test settings showed a significant performance gap between high-resource languages (English/French) and the African languages. Wolof is among the lowest-performing languages in the evaluation, mainly due to the small amount of publicly available data across the web (< 50 million characters) \citep{kudugunta2023} and its generally poor quality \citep{quality}. Furthermore, \texttt{given that low-resource languages are not all equally low in resources} \citep{nlplrlsurvey}, one phenomenon that remains understudied is the gap between African languages themselves. Across 42 supported African languages and 23 available public data sets, \citet{hussen2025} identified 04 languages (\texttt{Amharic}, \texttt{Swahili}, \texttt{Afrikaans}, and \texttt{Malagasy}) that are always treated, while there is over 98\% of unsupported African languages. This inequality also extends to the scripts used by these languages and can have various causes beyond the lack of data, such as \texttt{tokenization biases}, \texttt{high computational costs}, and \texttt{evaluation issues}. On this last point, \citet{jumelet2026multiblimp10massivelymultilingual} introduced \texttt{MultiBLiMP 1.0}, a massively multilingual morpho-syntactic evaluation benchmark covering 101 languages including Wolof, generated from Universal Dependencies treebanks using a controlled minimal-pair paradigm. Focused on subject-verb agreement, case assignment, and determiner-noun agreement, MultiBLiMP provides a fine-grained diagnostic tool for probing grammatical competence in LLMs. Evaluation of 20+ language models reveals that current LLMs perform poorly on morphologically complex, low-resource languages such as Wolof, exposing systematic gaps in grammatical understanding that aggregate benchmarks tend to mask. Addressing performance rather than evaluation, \citet{uemura2026merlinmultistagecurriculumalignment} proposed \texttt{MERLIN}, a multi-stage curriculum alignment framework that progressively aligns a multilingual text encoder with a generative LLM backbone. By combining encoder-side representation learning with instruction tuning and preference optimization, MERLIN achieves a +12.9 percentage point improvement on the AfriMGSM mathematical reasoning benchmark for African languages, illustrating how structured alignment pipelines can meaningfully narrow the performance gap for low-resource African languages in LLM evaluation settings.
Although tedious, costly, and time-consuming, data collection remains nevertheless a major stake in improving the representativeness of African languages on the global AI map. Therefore, \citep{afriwoz} introduced the first high-quality dialogue datasets for six African languages: Swahili, \texttt{Wolof}, Hausa, Nigerian Pidgin English, Kinyarwanda, and Yorùbá. The corpus consists of 1,500 turns each, which has been translated from a portion of the English Multi-Domain MultiWOZ dataset \citep{multiwoz} to enable the creation of dialogue agents for African languages. \textsc{AfriQA}, the first cross-lingual Question Answering (QA) dataset with a focus on African languages has been proposed in \citep{afriqa} laying the foundation for research on QA systems for one of the most linguistically diverse regions in the world. \textsc{AfriQA} includes +12,000 XOR QA examples across 10 African languages including Wolof. A \texttt{multiple-choice machine reading comprehension (MRC)} dataset spanning 122 language variants including Wolof, has been introduced in \citep{belebele}. Built from parallel passages in \texttt{FLORES-200} \citep{nllbteam2022}, the dataset enables controlled cross-lingual evaluation and reveals that, despite strong cross-lingual transfer in English-centric LLMs, smaller multilingual masked language models trained on balanced data exhibit broader language understanding. However, the best strategy to decide which cross-lingual data to include during training still remains an open question. Authors in \citep{blaschke-etal} analyzed transfer strategies between \texttt{263 different languages} (including Wolof), from \texttt{33 language families} across \texttt{03 tasks}: \textbf{POS tagging}, \textbf{Dependency parsing}, and \textbf{Topic classification}. They showed that beyond the definition of the concept of linguistic similarity, its effect on transfer performance depends mainly on the \textbf{NLP task} at hand, and the \textbf{(mono- or multilingual) input representations}. Regarding the latter, the authors in \citep{idris2026embedding} evaluated \texttt{05 embedding similarity metrics} across \texttt{03 multilingual models trained on African languages} (including Wolof). The actual transfer performance has been then measured on 03 downstream tasks: \textbf{NER}, \textbf{PoS-tagging}, and \textbf{Sentiment Analysis} for a total of \texttt{816 transfer experiments}. The eExperiments showed promising transfer prediction capabilities for NER and POS, with comparable predictive power to \texttt{URIEL} linguistic typology. An emerging direction that directly addresses the information access gap faced by primarily oral languages is cross-lingual speech-text retrieval. \citet{crosslingualmatryoshka} trained the first bilingual speech-text \texttt{Matryoshka} embedding model for Wolof and French, enabling direct retrieval of French documents from Wolof speech queries without relying on costly ASR-translation pipelines. Built on 860 hours of curated Wolof speech, the model demonstrates that a Late-Fusion architecture (HuBERT fused into a Matryoshka embedding LLM) generalises well beyond retrieval to tasks such as speech intent detection, illustrating the potential of cross-modal embeddings as a practical pathway to information access for speakers of oral-first languages.

Beyond cross-lingual transfer, LLMs also need to be fine-tuned on \texttt{instruction tuning data}\footnote{Input-output pairs that teach AI models to follow instructions, with each data point typically containing an \texttt{instruction}, an \texttt{input query}, and the \texttt{expected output response}.}, in order to make them more engaging in conversations, and enable them to properly follow the given instructions. This is particularly challenging given that this type of data is much more expensive to collect, as it requires human instructions and annotations \citep{ouyang2022}. \textsc{CohereLabs}\footnote{\url{https://cohere.com/research}} responded to this challenge by launching the \textsc{AYA} project: a year-long participatory research initiative that brought together nearly 3,000 participants from over 100 different countries \citep{singh-etal-2024-aya}. This made it possible to collect the largest native speaker instruction dataset with a total of 204K human-curated prompt-response pairs written by native speakers in 65 languages. Wolof is one of the languages that was not included in the initial list but was subsequently added thanks to the involvement of the \textsc{GalsenAI} community\footnote{Senegalese Artificial Intelligence community with thousands of members across Senegal, Africa, and the diaspora.} \citep{aya-at-a-glance}. This shows the importance of local AI communities in shaping African AI ecosystems, and GalsenAI is one of the pioneers that has played a major role in this regard in Senegal \citep{HENG2022102454}. Despite these efforts, sufficient data had not been collected in time, which led to the exclusion of Wolof from the training of the ensuing AYA models \citep{ustun2024}. A similar situation also happened during the \texttt{AfriqueLLM} project, whereby all languages for which a minimum of \texttt{90 million} tokens could not be collected (e.g. Wolof), were excluded from the training corpus \citep{afriquellm}. This highlights the importance of creating sufficiently large and clean pretraining corpora in order to improve the representativeness of national languages in LLM projects.
Historically, the first large-scale open-access language model to include Wolof in its training corpus was \texttt{BLOOM} \citep{workshop2023bloom}, a 176-billion-parameter multilingual causal language model jointly developed by over 1,000 researchers across 70 countries under the \textsc{BigScience} initiative. Trained on the \texttt{ROOTS} corpus; a carefully curated 1.6 terabyte multilingual dataset spanning 46 natural languages including Wolof and 13 programming languages \citep{NEURIPS2022_ce9e92e3}; BLOOM represents a historical milestone for the inclusion of Senegalese languages in large-scale AI systems. As the first truly open large language model, it established the foundation upon which subsequent African-focused multilingual models would build. In this regard, \citet{kudugunta2023} introduced \texttt{MADLAD-400}, a manually audited, general domain 3T token monolingual dataset based on \texttt{CommonCrawl}\footnote{\url{https://commoncrawl.org/}}, spanning 419 languages including Wolof and Pulaar. The authors discussed the limitations revealed by self-auditing MADLAD-400 (with poor quality Wolof crawled data), and the role data auditing had in the dataset creation process. To overcome this quality issue, \citet{penedo2025} introduced a new pre-training dataset curation pipeline based on \texttt{FineWeb} \citep{penedo2024} \textbf{that can be automatically adapted to support any language}. They then leveraged this pipeline to create \texttt{FineWeb2}, a new 20 terabyte (5 billion document) dataset covering over 1000 languages including Wolof and Pulaar. They also showed that models trained on language-specific corpora \texttt{produced by this pipeline}, perform better than those trained on other public web-based multilingual datasets. Building directly on this resource, \citet{fineweb2-hq-wolof} released \texttt{FineWeb2-HQ-50k-Wolof}, a synthetic Wolof pretraining corpus of 51,000 high-quality documents produced by machine-translating a curated subset of FineWeb2 into Wolof using the \textsc{Oolel} model \citep{oolel}. This initiative illustrates how machine translation, when combined with a high-quality multilingual source corpus, can serve as a scalable and cost-effective lever for bootstrapping pretraining data in extremely low-resource language settings. \citet{dossou-etal-2022-afrolm} introduced a novel self-active learning framework to pre-train a Language Model from scratch on 23 African languages including Wolof named \textsc{AfroLM}. With a dataset \texttt{14x smaller} than existing baselines, \textsc{AfroLM} outperforms many multilingual pretrained language models like AfriBERTa \citep{ogueji-etal-2021-small}, XLMR-base \citep{conneau-etal-2020} and mBERT \citep{devlin-etal-2019-bert}; on various NLP downstream tasks such as \texttt{Named Entity Recognition}, \texttt{Text Classification}, and \texttt{Sentiment Analysis}. To further enhance African language coverage in language models, researchers in \citep{adebara-etal-2023-serengeti} introduced \texttt{SERENGETI}, a set of massively multilingual language model that covers 517 African languages and language varieties including Wolof and Pulaar. They evaluated the models on 08 natural language understanding tasks across 20 datasets, and showed through meaningful comparisons, how SERENGETI model excels and acquire new SOTA. Extending this work to natural language generation, \citet{adebara-etal-2024-cheetah} introduced \texttt{Cheetah}, an encoder-decoder model pretrained on 517 African languages and fine-tuned on six NLG tasks including abstractive summarization, question generation, and machine translation. Cheetah significantly outperforms strong multilingual baselines such as mT5 and mBART across all tasks, establishing new state-of-the-art results for NLG in African languages. A complementary investigation into the monolingual versus multilingual trade-off has been presented in \citep{chang2026goldfishmonolinguallanguagemodels}, where the authors introduced \texttt{Goldfish}, a suite of monolingual language models trained for 350 individual languages including Wolof, using the GlotCC corpus. Despite using far fewer parameters than their multilingual counterparts, Goldfish models consistently outperform large multilingual models on perplexity and downstream tasks in the target language, demonstrating that dedicated monolingual pretraining remains a powerful strategy even in very low-resource settings. At a larger scale, \citet{salamanca2026tinyayabridgingscale} introduced \texttt{Tiny Aya}, a 3.35-billion-parameter generative model covering 70 languages with a strong focus on African ones, built on the Cohere \texttt{Command R} architecture. Fine-tuned on the multilingual AYA Collection and instruction-tuned using DPO alignment, Tiny Aya achieves state-of-the-art performance among open-source models of comparable size on multilingual generation benchmarks, making it one of the most capable small-scale multilingual LLMs currently available for African language tasks. The trajectory across this section reveals a recurring pattern: Wolof gains inclusion in continental and global benchmarks through community advocacy (GalsenAI in AYA) and community-driven data curation, yet consistently sits at the lowest-performing end of evaluations due to the scarcity and low quality of web-crawled data. The primary leverage point for improving Wolof LLM performance is therefore not architectural innovation but targeted data creation and curation.

In spite of all these challenges, conversational agents represent nevertheless a strategic lever for socioeconomic development, especially in the Senegalese context. Integrating automatic speech and language processing into systems adapted to local realities could significantly improve access to digital services in key sectors such as:

\begin{itemize}
    \item \textbf{Commerce:} Automation of customer services, intelligent recommendations, optimization of sales processes ;
    
    \item \textbf{Healthcare:} Digital medical assistance, easier access to health information, AI-assisted preliminary diagnosis ;

    \item \textbf{Banking and fintech:} Simplification of transactions, user support for mobile banking services ;
    
    \item \textbf{Education:} Access to online learning in local languages, interactive tutorials via educational chatbots, support for digital literacy.
\end{itemize}

The rise of dialogue systems in African languages is therefore a major opportunity to promote digital inclusion and ensure equitable access to information and communication technologies on the continent. This is particularly evident in the release of \textsc{AWA}, introduced as the first AI assistant capable of conversing fluently in Wolof \citep{awa}. Its announcement sparked unprecedented enthusiasm, highlighting the potential of such systems to have a direct and lasting impact on citizens' lives. However, proof of concept for such a Wolof conversational agent was explored in \citep{gauthier-etal-2022-preuve} well before the advent of AWA. Researchers opted for a modular architecture\footnote{Approach, in which the dialog system is broken down into a pipeline of different sub-tasks.} to design a conversational agent that provides information to the customers of a telecommunications provider. This approach is very common in the design of Task-oriented Dialogue Systems \citep{razumovskaia2022} and has long been used in the majority of chatbots deployed in industry prior to the recent rise of LLMs \citep{kang-etal-2018-data}. To overcome the limitations associated with manual collection of synthetic data in \citep{gauthier-etal-2022-preuve}, \citep{mbaye-diallo-2025-task} proposed a more scalable approach based on the \texttt{translate-train} paradigm. It is a general training approach to multilingual tasks where the key idea is to use the translator of the target language to generate training data in order to mitigate the gap between the source and target languages \citep{yu-etal-2022-translate}. Researchers proposed thus a chatbot generation engine based on the Rasa framework \citep{bocklisch2017} and a robust methodology for projecting annotations onto the Wolof language using an in-house machine translation system. Researchers in \citep{vwd} proposed an educational intelligent chatbot to improve literacy regarding the Von Willebrand disease (VWD)\footnote{A lifelong bleeding condition that makes it hard for blood to clot.} in Senegal. Their system is also based on the modular architecture with an \texttt{Automatic Speech Recognition (ASR)} system that converts spoken inputs into text which is then processed through the \texttt{Natural Language Understanding (NLU)} module\footnote{Authors used \href{https://www.certainly.io/}{\texttt{certainly}}, a proprietary conversational AI platform to train an NLU model.} to identify user intent across VWD-related themes. The chatbot generates appropriates responses via pre-defined templates containing culturally relevant information, and continuously improves its accuracy through a feedback loop that analyzes user interaction. \texttt{Baamtu Technologies} took a similar approach when setting up the SaytuTension chatbot, which aims to raise awareness about high blood pressure, a very common condition in Senegal \citep{IntraHealth2023}. Following the same design philosophy, \citep{chatbot-haemophilia} presented \textsc{Saytù Hemophilie}, a bilingual French-Wolof chatbot for haemophilia awareness in Senegal. The system follows a modular pipeline (ASR → NLU → response generation) and integrates both text and voice interaction modes. Evaluated with real users including patients, caregivers, and medical staff, the chatbot achieved a \texttt{System Usability Scale (SUS)} score of \textbf{81.7}, placing it in the ``Good'' usability category and demonstrating the practical viability of disease-specific multilingual conversational agents for health communication in Senegalese languages. These initiatives demonstrate a growing interest in designing practical applications based on these technologies for the common good.

Although more data-efficient than LLMs, this modular approach nevertheless suffers from \texttt{component isolation} i.e. each module must be optimized separately, and \texttt{error cascading}\footnote{Series of failures in a system of interconnected parts, where the failure of one component triggers the failure of others.} which make the maintenance of the overall system very tedious \citep{razumovskaia2022}. Systems such as AWA remain therefore quite promising for the local language inclusion, offering better interaction fluency and better scalability. Since AWA is a closed system, an \texttt{open source} alternative named \textsc{Oolel} and based on \textsc{Qwen 2.5} \citep{qwen2025}, was subsequently released in \citep{oolel}. It is one of the very first open source language models in Wolof that combines state-of-the-art AI technology with deep Wolof linguistic expertise.

\subsection{Speech Processing}
Speech processing is a field that analyzes and manipulates human speech signals using digital technology \citep{mehrish2023}. It involves a range of tasks like \texttt{Automatic Speech Recognition (ASR)} (converting speech to text), \texttt{Speech Synthesis} (text-to-speech), and \texttt{Speaker recognition}, and is used in applications such as \texttt{voice assistants} and transcription services. As with other NLP tasks, speech processing has also been revolutionized in recent years by deep learning approaches. These approaches are highly data-intensive, and much work has focused on data collection and more efficient data approaches.

\subsubsection{Automatic Speech Recognition}\label{sec:asr}
Automatic transcription of speech by any speaker in any environment is still far from solved, but ASR technology has matured to the point where it is now viable for many practical tasks \citep{jurafsky}. \citet{tamgno2011wolof} conducted a theoretical study in the areas of speech recognition in the Wolof language and presented the results obtained from their implementation with the \texttt{Julius} Speech Recognition \citep{akinobu_lee} open source software. Researchers leveraged an approach based on Hidden Markov Models (HMM) \citep{kakade2024} with a language model that consists of a word pronunciation dictionary and a syntactic constraint. In a favourable context for the development of a market for voice technologies in African languages, the \texttt{ALFFA} project has been launched in \citep{besacier:hal-01170505} to conduct fundamentals research on speech analysis (language phonetic and linguistic description, dialectology) and develop efficient speech technologies (ASR and TTS) for African languages. Authors described their achievements after 18 months of project and presented a multilingual calculator prototype in several African languages (including Wolof, Hausa and accented French) leveraging \texttt{Kaldi} \citep{Povey2011TheKS} for the ASR and the proprietary \texttt{Voxygen} \footnote{\url{https://www.voxygen.fr/home}} engine for the TTS. To cope with the vowel length contrast issue\footnote{Two versions (short/ long) of a same vowel that exist in the phoneme inventory of the language.} in languages like Wolof and Hausa, \citep{GAUTHIER2016136}, proposed a vowel length contrast modeling with length-contrasted and non-length-contrasted CD-DNN-HMM\footnote{Gaussian Mixture Model, Deep Neural Networks and Hidden Markov model.} models for ASR. They also used the Kaldi toolkit to train the Wolof model on 18,000 recorded utterances representing 21.3 hours of signal, and thus introduced the first large vocabulary continuous speech recognition system ever developed for the Wolof language. This work has subsequently been expanded to address a wider range of challenges faced by these languages, such as the \texttt{small amount of transcribed speech}, \texttt{written language normalization issues}, \texttt{limited text resources available for language modeling}, as well as \texttt{specific features (tones, morphology, etc.)}. Therefore, in addition to vowel length contrast modeling, data augmentation techniques through speed perturbation have been explored in \citep{gauthier:hal-01350057} to overcome the lack of resources. Researchers also developed ASR systems for Hausa and Wolof and made them openly available to the research community. They then leveraged the new research opportunities brought by growing digital archives and improving text and speech algorithms, to further explore automatic analysis approaches of the vowel length contrast phenomenon in \citep{gauthier2017}. Authors introduced multiple features to make a fine evaluation of the degree of length contrast under different factors and showed their abilities to properly highlight a variety of contrast degrees for each vowel considered. Still in the direction of data augmentation, \citet{synthetic} presented the first systematic assessment of large-scale synthetic voice corpora for African ASR. Researchers showed that the \texttt{generated synthetic voice data could be created for less than 1\% of the cost of collecting real human data}, while holding potential to complement this human data in creating and improving ASR models for African languages.

The ALFFA project has been completed afterwards in \citep{gauthier-etal-2016} where the researchers presented the data collected and ASR systems developed for 04 sub-saharan african languages (Swahili, Hausa, Amharic and Wolof). They illustrated their methodology by making a focus on Wolof for which they designed one of the first ASR systems ever built in this language and trained on 18k recorded utterances representing more than 21h of signal. All data and scripts had been made available online\footnote{\url{https://github.com/getalp/ALFFA_PUBLIC/tree/master/ASR/WOLOF}} and this dataset had a huge impact on the local NLP ecosystem, allowing a wide range of actors to experiment ASR systems in Wolof as in \citep{SB2021, jacobs2023}. At a larger scale, the \texttt{WAXAL} project \citep{diack2026waxal} released approximately 1,250 hours of image-prompted, transcribed ASR speech across 14 Sub-Saharan African languages, including Wolof and Fula. The data has been collected through a multi-year community partnership between \texttt{Google Research} and four African academic institutions (Makerere University, University of Ghana, Digital Umuganda, and Media Trust), providing one of the most comprehensive open speech resources for the region. However, when it was first released, the dataset had a major misalignment issue with the Wolof ASR data, rendering it unusable \citep{google-waxal-issues}. This was subsequently corrected by the GalsenAI community \citep{galsenai-waxal}: fixing broken training data that a trillion-dollar company shipped. This shows why open source matters, when the community catches what the corporation missed. \citet{asr-sota} conducted an extensive review of the state of the art in speech recognition in order to offer a comprehensive overview of the most recent and relevant developments in speech recognition. Researchers explored technological advances, cutting-edge algorithmic models, deep learning methodologies, and persistent challenges that drive research such as \texttt{low-resource languages}, multilingual models and innovation in this constantly evolving field. Different approaches that address the low-resource property of African languages in speech recognition have also been studied in \citep{AbdouMohamed}. The authors leveraged \texttt{self-supervised multilingual pretrained models} to introduce monolingual baselines and multilingual systems that are evaluated across 07 African languages including Wolof. They explored several multilingual strategies in order to mitigate language confusion and lexical ambiguity, and demonstrated that incorporating language-aware mechanisms, improves multilingual ASR performance while reducing reliance on external language identification. To help build foundational digital resources for African languages, the \texttt{AI4D - African Language Program}\footnote{\url{https://k4all.org/project/language-dataset-fellowship/}} has been launched with 03 main objectives:

\begin{itemize}
    \item \textbf{Incentivise} the crowd-sourcing, collection and curation of language datasets through an online quantitative and qualitative challenge ;
    \item \textbf{Support} research fellows for a period of 3-4 months to create datasets annotated for NLP tasks ;
    \item \textbf{Host} competitive Machine Learning challenges on the basis of these datasets.
\end{itemize}

It is within this context that Baamtu Technologies launched the \texttt{AI4D Baamtu Datamation Automatic Speech Recognition in WOLOF} hackathon\footnote{\url{https://zindi.africa/competitions/ai4d-baamtu-datamation-automatic-speech-recognition-in-wolof}} bringing together more than 200 participants and leading to the design of an ASR model achieving a \texttt{Word Error Rate} (WER) of \texttt{0.110} \citep{siminyu2021}. This initiative has strengthened interest in the ecosystem and reinforced Baamtu Technologies' position as a local leader in NLP for local languages, particularly with its \texttt{Kàllaama} suite\footnote{\url{https://www.youtube.com/watch?v=P5PRgugOu8o}}. Similarly to machine translation, tech giants are also interested in speech processing in under-resourced languages. \texttt{Meta} (formerly Facebook) has notably launched the \texttt{Massively Multilingual Speech (MMS)} project in \citep{pratap2023} that increases the number of supported languages by 10-40x, depending on the task. The core of this work involved creating a new dataset from readings of publicly available religious texts and effectively using \texttt{self-supervised learning} \citep{Ericsson2022}. This effort resulted in pre-trained \texttt{wav2vec 2.0} models \citep{baevski2020} for thousands of languages, a single multilingual speech recognition and synthesis models each covering 1,107 languages, as well as a language identification model for more than 4k languages.
They recently introduced \texttt{Omnilingual ASR} \citep{omnilingualasr}, the first large-scale ASR system designed for extensibility, which allows communities to introduce unserved languages \texttt{with only a handful of data samples}. This model scales self-supervised pre-training to 7B parameters to learn robust speech representations, and introduces an encoder-decoder architecture designed for zero-shot generalization, leveraging a LLM-inspired decoder. Omnilingual ASR expands coverage to over 1,600 languages including Wolof (Senegalese and Gambian) and Fula, the largest such effort to date, including over 500 never before served by ASR. This kind of initiatives drastically reduce the entry barrier to developing advanced language processing models in these languages, which encourages local players such as \texttt{LAfricaMobile}\footnote{\url{https://lafricamobile.com/en/produit-stt/}} and \texttt{Lengo} \citep{lengo} to offer Wolof ASR for different use cases. Even \texttt{Orange}\footnote{Leading telecom operator in Senegal.}, which was relatively inactive at the beginning of the local AI ecosystem's emergence \citep{HENG2022102454}, is now taking a keen interest in local languages. They recently announced their partnership with \texttt{OpenAI} and \texttt{Meta}, to fine-tune Large Language models (LLMs) to understand regional languages in Africa that today are not understood by any GenAI model, with an initial focus on Wolof and Pulaar \citep{orangegenai}. This follows the advent of \texttt{Whisper} models \citep{radford2022} developed by OpenAI, which show promising performance when fine-tuned on Wolof data as in \citep{yux}. The researchers focused on \texttt{Maternal and Reproductive Health} and collected 250 essential healthcare keywords that has been expanded to 750 real-world phrases, and translated them into Wolof and Hausa. They used a Whisper model initially fine-tuned on Wolof \citep{whosper2025} to adapt it to the medical domain via the \texttt{LoRA (Low-Rank Adaptation)} approach \citep{hu2021}, which requires fewer computational resources. This is particularly relevant as domains such as \textbf{healthcare} typically suffer from a \texttt{double resource scarcity}, where there is both a lack of language and domain data \citep{nlplrlsurvey}.

Beyond proprietary projects, \texttt{Orange} is also active in the open source community with the introduction of the first self-supervised multilingual speech model trained exclusively on African speech \citep{caubrière2024}. They pre-trained an \texttt{HuBERT-based} model \citep{hsu2021hubertself} on nearly 60k hours of unlabeled speech segments, in 21 languages and dialects spoken in sub-Saharan Africa (including Wolof and Pulaar) \citep{caubriere-gauthier}. The evaluation on the SSA subset of the \texttt{FLEURS-102} dataset \citep{conneau2022fleurs} demonstrated competitive results. This confirms the findings in \citep{djionangpindoh} which showed that combining under-resourced languages that share similar linguistic and phonetic characteristics (Wolof, Swahili and Fongbe), enhances the quality of features extracted for each language individually. \citet{sy2025speech} went further by curating around 1.4 TB of raw Wolof speech and filtered it down to 860 hours of high-quality spontaneous audio, using a multi-stage pipeline (source separation, diarization, VAD, quality filtering). They performed continued pretraining \citep{parmar2024reuse} of HuBERT on this Wolof data, which allows to improve ASR performance over both the original Meta/Hubert model \citep{hsu2021hubertself} and Orange/HuBERT \citep{caubrière2024}, \texttt{while using far less language-specific compute}. \citet{kandji2025wolbanking77} introduced the first Wolof spoken intent classification dataset consisting of more than 04 hours of spoken customer service queries. Researchers fine-tuned multilingual pre-trained ASR models and conducted extensive evaluations. Although the dataset remains small, it nevertheless constitutes the first initiative paving the way for research in \texttt{Spoken Language Understanding} in local languages (cf Section \ref{sds}). Going beyond recognition, \citet{crosslingualmatryoshka} leveraged the same 860 hours of curated Wolof speech \citep{sy2025speech} to train the first cross-lingual speech-text \texttt{Matryoshka} embedding model for Wolof-French, enabling direct retrieval of French documents from Wolof speech queries and avoiding the error propagation inherent in cascaded ASR-translation pipelines. At a larger scale, \citet{omnilingualsonarteam} extended the \texttt{OmniSONAR} sentence embedding space to the speech modality by mapping 177 spoken languages into a shared semantic space via teacher-student distillation on ASR data, achieving a 43\% lower error rate in cross-lingual and cross-modal similarity search and 97\% of SeamlessM4T performance on speech-to-text translation in a zero-shot setting. At the continental scale, \citet{elmadany-etal-2025-voice} undertook a systematic mapping of Africa's speech technology landscape, introducing \texttt{SimbaBench}, a unified evaluation framework aggregating all publicly available ASR, TTS, and spoken language identification (SLID) resources across 61 African languages including Wolof. Using SimbaBench, the authors trained the \texttt{Simba} family of fine-tuned models and established new state-of-the-art performance across multiple African languages and speech tasks, while revealing critical patterns in resource availability and the influence of domain diversity and language family relationships on model performance. Complementing this broad coverage, \citet{sunbird2026asr51} released \texttt{asr-whisper-51-african-languages}, a Whisper Large v3 model fine-tuned on 51 African languages including Wolof, built on aggregated community datasets spanning Google FLEURS \citep{conneau2022fleurs}, the ASR Africa Efficiency Benchmark, and Sunbird SALT. From an evaluation infrastructure standpoint, \citet{muchai2026paza} introduced \texttt{PazaBench}, the first ASR leaderboard dedicated to low-resource languages, initially covering 39 African languages and benchmarking 52 state-of-the-art ASR models across three standardized metrics: CER, WER, and RTFx (real-time factor). PazaBench promotes a human-centered evaluation methodology grounded in real-world community testing, and highlights where African-centric specialized models outperform generic large multilingual systems. On the commercial side, \citet{elevenlabs2026scribe} released \texttt{Scribe v1}, a proprietary ASR system covering 99 languages including Wolof, achieving a WER of \textbf{40.7\%} on the FLEURS benchmark for Wolof; outperforming Whisper Large v3 (97.4\%), Gemini Flash 2 (48.4\%), and Deepgram Nova 2 (100\%). At the local level, \texttt{AI Hub Senegal} introduced two Whisper-based systems: \texttt{Kiriku-Wolof-ASR}, a 2-billion-parameter ASR model for Wolof built on community audio data \citep{aihubsn2026kiriku-wolof-asr}, and \texttt{M-Kiriku ASR}, the first open-source multilingual ASR model explicitly covering Wolof, Pulaar, and Sérère \citep{iahubsn2026m-kiriku-asr}. Trained on 150 hours of curated multilingual speech spanning agriculture, banking, urban life, and daily conversation, M-Kiriku ASR extends the Whisper Large v3 vocabulary with custom language tokens and language-specific characters, and achieves WERs of \textbf{8.56\%}, \textbf{24.45\%}, and \textbf{19.80\%} for Wolof, Pulaar, and Sérère respectively. The spread of WERs across published systems on the FLEURS-Wolof benchmark is extreme; from Whisper Large v3 at 97.4\% to M-Kiriku at 8.56\%; and this spread is explained by three compounding factors rather than model architecture alone. First, \textbf{language-specific fine-tuning depth}: zero-shot multilingual models degrade sharply on Wolof because their pretraining corpus contains very little Wolof data \citep{kudugunta2023}, while systems fine-tuned directly on Wolof speech (Kiriku, Sy et al. HuBERT) benefit from in-domain feature adaptation. Second, \textbf{vocabulary and tokenization}: M-Kiriku's explicit extension of the Whisper vocabulary with Wolof-specific characters avoids systematic subword fragmentation errors that inflate the WER of standard tokenizers on words with Wolof diacritics. Third, \textbf{training data domain}: models trained on conversational and spontaneous Wolof speech \citep{sy2025speech} generalize better to the FLEURS evaluation style than models trained on read-speech or religious-text corpora. These factors also explain why ElevenLabs Scribe (40.7\% WER), despite being a large proprietary system, performs worse than the locally fine-tuned M-Kiriku: scale without language-specific adaptation is insufficient for Wolof ASR. The ASR literature for Senegalese languages traces three successive paradigm shifts: HMM-based systems trained on tens of hours \citep{GAUTHIER2016136}, self-supervised pre-training on unlabeled speech at the tens-of-thousands-of-hours scale (MMS, HuBERT-Orange), and LLM-inspired encoder-decoder architectures (Omnilingual ASR). Throughout these transitions, community-driven correction and local fine-tuning; exemplified by GalsenAI's repair of the WAXAL dataset and the Kiriku initiative; have consistently delivered the strongest results for Senegalese languages, while large multilingual systems provide a practical starting point but not a final solution. Wolof now has a credible open ASR pipeline; extending comparable coverage to Pulaar, Sérère, and the remaining three national languages is the next critical frontier.

\subsubsection{Speech Synthesis}
The modern task of text-to-speech or TTS, also called speech synthesis, is exactly the reverse of ASR: \texttt{to map text to an acoustic waveform} \citep{jurafsky}. It is used in applications like spoken language models that interact with people or for reading text out loud, greatly improving inclusion in information access. \citet{tamgno2011wolof} conducted a theoretical study in Wolof speech synthesis and implemented a basic TTS system with the festival speech tool \citep{festival2014manual}. Beyond models, authors also created different lexicons and knowledge bases of phonetic, acoustic and linguistic features in order to introduce other languages. As part of the \texttt{Cracking the Language Barrier for a Multilingual Africa} project\footnote{\url{https://k4all.org/project/building-wolof-text-to-speech-system/}}, the first open Wolof TTS dataset has been introduced in \citep{woltts}. It contains recordings from two native Wolof voice actors (a male and female voice). Each actor recorded more than 20,000 sentences for approximately 19 hours of recordings for the female voice and 22 hours for the male voice. Having identified artifacts in the textual data as well as poor recording quality, the GalsenAI community proposed a cleaned version of this dataset in \citep{galsenai_anta}. They extracted the \texttt{female voice}, denoised it and enhanced it with the \texttt{Resemble Enhance} library \citep{resemble_enhance_2023} and removed emojis, special characters as well as Arabic and Russian characters. They also removed audios judged not qualitative enough, reducing its size to around 18h40mn of high-quality recordings. They subsequently trained a TTS model based on \texttt{xTTS-V2} \citep{casanova2024} on the cleaned dataset and openly released it for public use \citep{galsenai_tts}. This initiative has inspired other local stakeholders such as \texttt{Concree}\footnote{\url{https://concree.com/}}, which introduced \texttt{AdiaTTS} \citep{concree} based on the \texttt{ParlerTTS} model \citep{lacombe-etal,lyth2024natural}. Their model is trained on 40 hours of Wolof speech data, which has not been published. These initiatives are generating real enthusiasm for Wolof speech synthesis systems, and the 2025 CNRIA\footnote{Conference on Research in Computer Science and its Applications} Demo Paper Award was even given to a demo on this topic \citep{tts-wolof}. Given the difficulty of accessing high-quality data for languages with limited resources, \citet{ogayo2022} explored efficient corpus creation and sharing approaches as well as the deployment of TTS systems. They also demonstrated it was possible to develop synthesizers that generate intelligible speech with only 25 minutes of created speech, even when recorded in suboptimal environments. Speech data, code, and trained models for 12 African languages including Wolof was subsequently released for future research. Complementing these single-language datasets, the \texttt{WAXAL-TTS} corpus \citep{diack2026waxal} provides over 235 hours of studio-quality, single-speaker recordings across 13 Sub-Saharan African languages, including Wolof and Fula, making it the largest publicly available multilingual TTS dataset for African languages to date. Building TTS pipelines for these languages further requires robust Grapheme-to-Phoneme (G2P) preprocessing which the \texttt{africa-g2p} library \citep{michsethowusu2026africag2p} addresses. They released a rule-based system that converts text in over 400 African languages, including Wolof and Pulaar, into native-orthography phoneme units or IPA, directly usable as input for speech synthesis training. \texttt{LAfricaMobile}\footnote{\url{https://lafricamobile.com/en/produit-tts/}} offers the widest language coverage in its TTS in Senegal with Wolof, Bambara, Dioula, Lingala, Hausa, and Fulfulde. As a private company, all of its models and data are kept private. Locally, AI Hub Senegal extended its \texttt{Kiriku} initiative to speech synthesis by releasing two VITS-based TTS systems via the Coqui TTS framework: \texttt{Kiriku-Wolof-TTS}, a single-speaker end-to-end Wolof TTS model trained directly on graphemes with a 93-character vocabulary covering the Latin alphabet and Wolof-specific diacritics \citep{aihubsn2026kiriku-wolof-tts}, and \texttt{Kiriku-Pulaar-TTS}, the first open-source TTS system for the Pulaar language, using the same architecture \citep{aihubsn2026kiriku-pulaar-tts}. Both models jointly learn the text-to-spectrogram and spectrogram-to-waveform steps without a separate vocoder, enabling lightweight deployment. Adding expressive capabilities, \citet{oolel-voices} introduced \texttt{Oolel-Voice}, a 0.5-billion-parameter Wolof TTS model by Soynade Research that supports voice cloning with fine-grained control over tone, pace, and emotion, making it suitable for a wide range of applications from conversational AI and virtual assistants to dubbing, audiobook narration, and e-learning. At the global scale, \citet{zhu2026omnivoice} introduced \texttt{OmniVoice}, a massively multilingual zero-shot TTS system supporting over 600 languages including Wolof, Gambian Wolof, and Pulaar. Trained on a 581k-hour open-source multilingual dataset, OmniVoice achieves the broadest language coverage reported to date through a diffusion language model-style non-autoregressive architecture that directly maps text to multi-codebook acoustic tokens, and delivers state-of-the-art performance on multilingual TTS benchmarks covering up to 102 languages. TTS for Wolof has progressed faster than ASR in the recent period: multiple competing open-source systems now exist (GalsenAI xTTS-v2, AdiaTTS, Kiriku-Wolof-TTS, Oolel-Voice), a large studio-quality multilingual corpus (WAXAL-TTS) and a dedicated G2P library (africa-g2p) are both publicly available. By contrast, TTS systems for Pulaar, Sérère, and the other three national languages remain at the single-model proof-of-concept stage, and none address the domain or speaker diversity needed for practical deployment. The WAXAL-TTS and africa-g2p resources are the most actionable levers for closing this gap.

\subsubsection{Spoken Dialog Systems}\label{sds}
A spoken dialogue model refers to a dialogue system capable of generating intelligent verbal responses based on the input speech \citep{ji2024wavchat}. It has two essential components that do not exist in a written text dialog system: a \texttt{speech recognizer} and a \texttt{text-to-speech} module. Spoken Dialog Systems represent thus one of the most direct methods of \texttt{human-computer interaction (HCI)}\footnote{Study and design of how people and computers interact with each other, focusing on creating user-friendly and effective technology interfaces.}, and has tremendous potential in the African context, given that African languages are spoken more than they are written \citep{ailanggap}. The proof of concept presented in \citep{gauthier-etal-2022-preuve} and highlighted in Section \ref{dialog} is the first Wolof dialog system documented in research. Authors used an in-house speech recognition model trained on 132 hours of data (1/3 gold data, 2/3 synthetic data), which is based on \texttt{Kaldi} \citep{Povey2011TheKS}. The researchers used pre-recorded messages instead of a speech synthesis system to return the output to the user. The introduction of voice features also increases the complexity of the architecture by adding additional components, exacerbating the problem of cascading errors (cf Section \ref{dialog}). One way to address this issue is to replace the ASR and the NLU components by a single one, that directly maps the audio input to the corresponding intent (SLU) \citep{mesnil2014}. Therefore, \citep{kandji2025wolbanking77} introduced the first Wolof spoken intent classification dataset consisting of more than 4 hours of spoken customer service queries. Researchers explored speech recognition models and leave room for research in \texttt{spoken intent classification} in Wolof by openly releasing the dataset. In \citep{awa}, voice features are also presented as part of the AWA dialogue system, but no information was shared about the ASR and TTS used.

Another approach is to design an entirely end-to-end architecture, that directly processes the audio input to produce audio output. Although the definition of \texttt{Speech LLMs} remains non-standardized in current research \citep{peng2025}, this concept can be defined as \texttt{speech-to-speech generation tasks} \citep{arora2025}. In this way, \citep{sy2025speech} introduced the first speech language model for Wolof alongside with a 860 hours spontaneous and high-quality unsupervised speech dataset. Authors highlighted the effectiveness of continued pretraining \citep{parmar2024reuse}, which allows to reuse the compute already invested in the base model. The \textsc{GalsenAI} community introduced the first \texttt{Keyword spotting} (KWS)\footnote{The task of learning to detect spoken keywords.} dataset that covers all the 06 senegalese languages presented in this survey \citep{galsenai-keywords}. They extended the Speech commands dataset \citep{warden2018} that included a limited vocabulary composed of around twenty common words at its core i.e. digits from zero to nine, and seventeen words that would be useful as commands in \texttt{IoT} or \texttt{Robotics} applications. The development of keyword spotting models has been explored as part of a dedicated hackathon\footnote{\url{https://www.kaggle.com/competitions/indabaxsenegal2023-wolof-language-keyword-spotting}}, showcasing the project while training young people in the fundamentals of machine learning. Researchers in \citep{jacobs2023} proposed an alternative to the conventional KWS approach that involves transcribing the audio corpus with an automatic speech recognition (ASR) system and then searching for keywords in the output. They introduced an ASR-free approach by extending the \texttt{query-by-example (QbE)} methodology with \texttt{multilingual acoustic word embeddings (AWEs)} and compare their effectiveness w.r.t "classical" ASR-based methods. In controlled experiments on Wolof and Swahili where training and test data are from the same domain, their results showed that an ASR model trained on just five minutes of data outperforms the AWE approach.

While sharing some similarities with spoken dialog systems, command and control speech systems can be further distinguished from them, as they are able to respond to requests but do not attempt to maintain continuity over time. The spoken dialogue research reviewed here traces a clear architectural arc: modular pipelines (ASR → NLU → response generation) enabled the first Wolof dialogue prototypes but are limited by cascading errors; end-to-end and speech LLM approaches (Sy et al., AWA, Oolel) promise better fluency but require far larger resources. The keyword spotting and intent classification datasets introduced for all six national languages represent the community's most inclusive contribution to date, establishing a shared evaluation foundation for the speech understanding tasks.

\subsection*{Cross-Language and Cross-Task Coverage Summary}
\label{sec:coverage}
Table~\ref{tab:coverage} synthesizes the resource landscape described in Section~\ref{sec:lrcontext} by mapping NLP task coverage against the six national languages. A filled circle (\textbullet) indicates that at least one dedicated resource, tool, or model exists; a half-filled marker ($\circ$) denotes minimal or partial coverage (e.g., a single small corpus, or coverage as part of a multilingual dataset without Senegal-specific evaluation); a dash (---) indicates no published resource. The table makes the Wolof–others asymmetry immediately visible: Wolof has some form of coverage across all task categories, whereas the remaining five languages collectively hold full coverage only in keyword spotting, which is the sole task for which a six-language dataset was released simultaneously.

\begin{table*}[!htbp]
\centering
\caption{NLP task coverage by Senegalese national language. \textbullet{} = dedicated resource/model exists - $\circ$ = minimal or partial (e.g., included in a multilingual dataset without language-specific evaluation) - ; = no published resource.}
\label{tab:coverage}
\resizebox{\textwidth}{!}{%
\begin{tabular}{lcccccc}
\toprule
\textbf{NLP Task} & \textbf{Wolof} & \textbf{Pulaar} & \textbf{Sérère} & \textbf{Diola} & \textbf{Mandingue} & \textbf{Soninké} \\
\midrule
Language Identification (LID)          & \textbullet & $\circ$ & $\circ$ & $\circ$ & $\circ$ & $\circ$ \\
Parsing \& Tokenization                & \textbullet &;     &;     &;     &;     &;     \\
Named Entity Recognition (NER)         & \textbullet &;     &;     &;     &;     &;     \\
Part-of-Speech Tagging (POS)           & \textbullet &;     &;     &;     &;     &;     \\
Sentiment Analysis                     & \textbullet &;     &;     &;     &;     &;     \\
Hate Speech Detection                  & \textbullet &;     &;     &;     &;     &;     \\
Intent Classification                  & \textbullet &;     &;     &;     &;     &;     \\
Fact Checking                          & \textbullet &;     &;     &;     &;     &;     \\
Spell Checking                         & \textbullet &;     &;     &;     &;     &;     \\
Machine Translation (MT)               & \textbullet & $\circ$ &;     &;     &;     &;     \\
Automatic Speech Recognition (ASR)     & \textbullet & $\circ$ & $\circ$ &;     &;     &;     \\
Text-to-Speech Synthesis (TTS)         & \textbullet &;     &;     &;     &;     &;     \\
Keyword Spotting (KWS)                 & \textbullet & \textbullet & \textbullet & \textbullet & \textbullet & \textbullet \\
Spoken Dialogue Systems                & \textbullet &;     &;     &;     &;     &;     \\
\bottomrule
\end{tabular}%
}
\end{table*}

\section{Case Study: NLP Pipelines for the Social Sciences}
Natural Language Processing has immense potential to transform methodologies in social science, particularly in multilingual African contexts such as Senegal. NLP tools can reduce the time, cost, and cognitive burden of qualitative research while broadening access to linguistic data collected in national languages (cf Section \ref{sec:sentiment}). Field-based social research in Senegal typically relies on extensive interviews, focus groups, and participant observations conducted in multiple languages, often Wolof and French \citep{saintlouis, saintlouisbis}. Manual transcription and translation of these data are among the most time-consuming phases of the research process. To prevent this step from becoming an obstacle to data exploitation, researchers generally include the costs of transcribing interviews conducted in national languages in their budgets. Furthermore, to optimize efficiency, interviews conducted in a local language are transcribed by young researchers (doctoral students and linguistics students) who understand the language of the respondents. The challenges researchers encounter in transcribing survey data into local languages “forces” them to \texttt{favor thematic analysis over lexical analysis}, which requires manual transcription\footnote{Feedback shared by a small group of local social science researchers we interviewed.}. Integrating speech recognition (ASR) and machine translation (MT) systems into the workflow, can drastically reduce transcription time and costs, while preserving multilingual fidelity \citep{yang-etal-2022-jhu}. The interviews that have already been transcribed could have served as an interesting corpus for ASR training, but the personal data they contain makes it difficult to collect them. Automatic alignment and metadata tagging (cf Section \ref{tok-class}) can also further facilitate corpus analysis, making it easier to identify recurrent themes or sentiments across respondents and regions (cf Section \ref{sec:sentiment}). 

However, automating qualitative analysis introduces ethical and privacy challenges, particularly when dealing with sensitive or identifiable data. The study in \citep{nlpshs} offers a comprehensive synthesis of how NLP techniques are increasingly reshaping social science research. It outlines a three-layer framework, from \texttt{preprocessing} and \texttt{representing unstructured text}, to \texttt{extracting semantic information}, and finally \texttt{applying these insights to sociological and political analyses}. The review highlights how \texttt{classical methods} (e.g., dictionaries, topic models, supervised classifiers) and \texttt{modern deep-learning approaches} (e.g., contextual embeddings, transformers, large language models) enable researchers to work at unprecedented scale while uncovering complex patterns. These patterns can be related to bias, culture, political behavior, online polarization, and collective action. The authors also identify key methodological challenges, including representativeness, model bias, and interpretability. They also discussed future directions, especially the growing role of \texttt{large language models (LLMs)} in annotation, research design, and simulation. Although LLMs now play a central role in tasks such as large-scale text coding \citep{textcoding}, synthetic data generation \citep{Gilardi_2023}, and the simulation of human judgments \citep{chiang2023}, researchers in \citep{llmshs} argued that, their deployment raises unresolved concerns regarding reliability, validity, replicability, and model drift. Recent methodological work in computational social science has therefore emphasized the need for rigorous frameworks when using LLMs for empirical research \citep{suri2023}. To address these issues, the authors in \citep{llmshs} outlined a set of best practices, such as \texttt{transparent reporting of model specifications}, \texttt{systematic validation against human-coded ground truth}, and \texttt{explicit handling of nondeterminism}, that are essential to ensure credible scientific inference. 

These works are particularly relevant for research on low-resource languages such as Wolof, where methodological opacity or unvalidated model behavior could disproportionately distort scientific findings (cf Section \ref{sec:mt}). By foregrounding replicability and the careful evaluation of model outputs, this primer offers a transferable roadmap for responsibly integrating NLP tools into linguistic and socio-cultural analyses involving African languages.

\section{Discussions \& Perspectives}

\subsection{Challenges}\label{sec:challenges}
Despite growing momentum, the development of NLP for Senegalese national languages faces multiple interrelated challenges spanning data, methodology, infrastructure, and governance. The technical survey in Section~\ref{sec:effort} provides the evidential basis for each challenge identified here, and data scarcity has been documented across all seven NLP task clusters. Linguistic complexity surfaces prominently in tokenization and spell checking, while reproducibility gaps remain most acute in machine translation, where the French--Wolof parallel datasets remain unpublished and only one locally created parallel corpus, the Wolof-Arabic \texttt{MudawanSn} \citep{mbaye-diop2026gold}, has been publicly release. The uneven distribution of resources across the six national languages also runs as a thread through every subsection, with Wolof consistently the best-served and Joola, Mandinka, and Soninké the least.

\textbf{\textsc{Data Scarcity and Quality:}}
The most critical limitation remains the scarcity of high-quality, standardized linguistic data. Existing corpora are fragmented across institutions and often lack sufficient size, metadata, or representativeness. Many textual sources are informal or domain-specific (e.g., religious texts or social media posts), introducing stylistic bias. For speech, recordings are limited in both duration and speaker diversity, while annotation is slow due to the lack of trained linguists in the different languages and \textbf{funding}. Overall, the progress to date underscores both the rapid advances and the persistent inequalities across languages. While Wolof has benefited from relatively larger datasets and visibility in continental benchmarks, other national languages remain at the exploratory stage, illustrated in Table~\ref{tab:coverage}. Continued investment in data curation, open evaluation, and multi-institutional collaboration will be critical to ensure balanced development across Senegal’s linguistic spectrum.

\textbf{\textsc{Linguistic Complexity and Orthographic Variation:}}
The structural diversity of Senegal’s languages presents unique modeling challenges. Tonal contrasts, morphophonemic alternations, and agglutinative morphology affect tokenization and subword segmentation. In addition, orthographic conventions are still evolving and inconsistently encoded across digital platforms. The lack of standard transliteration systems further complicates data integration across dialects and scripts, and reinforces the dominance of Wolof over other local languages.

\textbf{\textsc{Limited Computational Infrastructure:}}
Sustained progress requires access to modern compute infrastructure and cloud services for model training and evaluation. While the government invested in a national compute infrastructure \citep{supercomputer}, the latter struggles to be made operational \citep{notworking} and most local universities still rely on limited or outdated hardware. Dependence on foreign hosting services raises additional issues of data sovereignty and sustainability.

\textbf{\textsc{Capacity Building and Skill Gaps:}}
The availability of trained researchers and engineers in NLP remains limited. A recent continental talent mapping study \citep{gsma2026talent} reveals that West Africa faces high-priority shortages across six of the seven core AI skill clusters, with Linguistic \& Cultural Expertise recording an availability score of only \texttt{2.0 out of 5.0}. This reflects a severely constrained practitioner pool despite the region's leading share of open-source AI repositories. Formal university programmes alone receive an average effectiveness rating of only \texttt{3.5 out of 5.0} for producing job-ready AI professionals, compared to \texttt{4.5} for technology innovation hubs and project-based learning environments. \textbf{Brain drain} further compounds these challenges, as the migration of skilled researchers toward better-funded international institutions, attracted by competitive doctoral fellowships and higher salaries, generates an oversupply of junior talent alongside a critical shortage of experienced mentors and technical leaders \citep{gsma2026talent}. Although institutions and local universities are expanding training programs, the scale of need outpaces current capacity. Bridging this gap demands long-term investment in curriculum development, mentorship, and community-based learning initiatives.

\begin{figure*}[!htbp]
\centering
\includegraphics[width=0.88\textwidth]{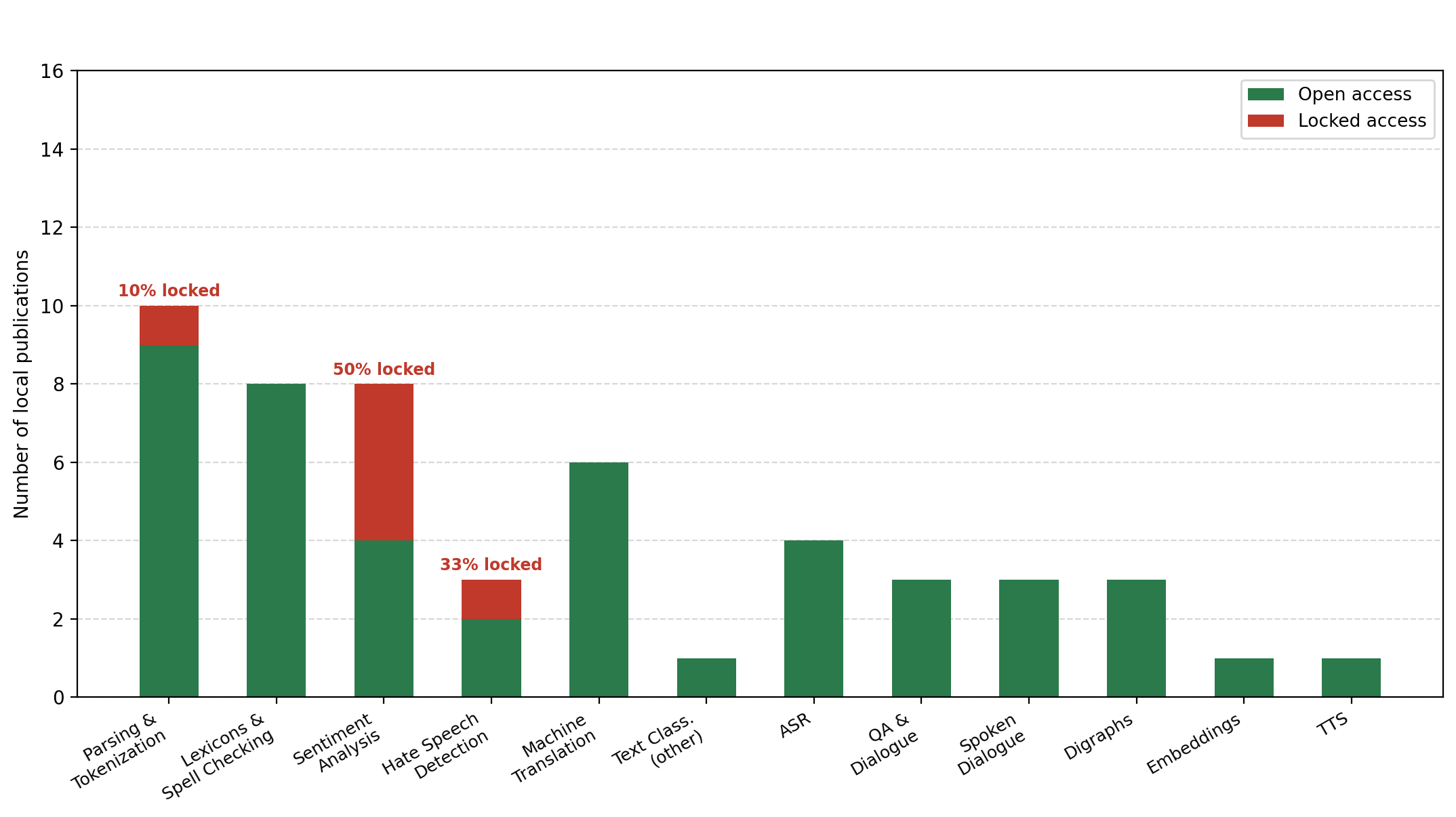}
\caption{Open-access versus locked-access distribution among local (Senegalese) peer-reviewed publications by NLP task. Tasks not shown have no locked local publications.}
\label{fig:locked_open}
\end{figure*}

\textbf{\textsc{Visibility of local research}}
Local research also lacks visibility, with the vast majority of articles not being published in major conferences. To date, \citep{kandji2025wolbanking77} is the only local research work that has been published at \textbf{NeurIPS}, which is one of the major conferences in Artificial Intelligence. Researchers would therefore benefit from targeting conferences and journals such as \texttt{NeurIPS}, \texttt{AAAI}, \texttt{ICLR}, \texttt{ICML}, \texttt{IJCAI}, \texttt{InterSpeech}, \texttt{EMNLP}, \texttt{ACL}, \texttt{TACL}, etc. Moreover, a lot of them are locked behind paywalls making it difficult for most researchers who do not have access licenses. As illustrated in Figure~\ref{fig:locked_open}, \texttt{Sentiment Analysis} stands out with 50\% of its papers behind a paywall, all published in \texttt{IEEE conference proceedings}. This concentration of paywalled work in the most active local NLP sub-field directly constrains reproducibility and community uptake.

\textbf{\textsc{Ethical, Legal, and Governance Challenges:}}
Ethical and governance frameworks for language data remain underdeveloped. Questions of consent, cultural ownership, and intellectual property are critical for corpora derived from oral traditions and social research. Ensuring compliance with both \texttt{CDP}\footnote{Senegalese Committee for the Protection of Sensitive Data.} and \texttt{SODAV}\footnote{Senegalese Agency for Copyright and Related Rights.} principles requires institutional coordination. Without such alignment, the risk of data misuse or extractive research practices persists. Therefore, the \texttt{UNESCO AI Readiness Assessment Methodology (RAM)} report \citep{unesco} highlighted the urgent need for a strong, centralized AI and data governance body, with technical sectoral committees and ethical oversight.

\subsection{Opportunities and Future Directions}
While Senegal’s AI ecosystem remains nascent, the convergence of \texttt{political will} \citep{aistrategy, aieducation, aiinschool, gates}, \texttt{community engagement} \citep{senhubia, indaba2024, saltis}, and \texttt{academic expertise} \citep{ailab} provides fertile ground for sustainable growth. This section outlines several key opportunities and future directions that could accelerate progress across linguistic, technical, and social dimensions.

\textbf{\textsc{Linguistically-Informed and Typology-Aware Development:}} A central recommendation drawn from \citet{adebara-abdul-mageed-2022-towards} is that NLP systems for African languages must move beyond Indo-Eurocentric algorithm design assumptions and actively exploit the typological properties of the target languages. For Senegalese national languages, three features are particularly underexplored. \textbf{First, tone} is a grammatically functional property in Pulaar, Mandinka, and Soninké: stripping tone diacritics during preprocessing introduces lexical and syntactic ambiguities that degrade POS tagging, text classification, and machine translation quality, while ASR and TTS systems for these languages must be explicitly designed to handle tonal contrasts at both the acoustic and orthographic levels. \textbf{Second, vowel harmony} in Wolof constrains the set of permissible vowel co-occurrences within a word, a regularity that can be actively leveraged by spell checkers (as a phonological plausibility filter) and POS taggers (tokens sharing the same grammatical category tend to exhibit similar harmonic patterns). \textbf{Third, serial verb constructions (SVCs)} in Mandinka; in which multiple verbs combine without conjunctions; are poorly handled by standard word-embedding and MT architectures that treat each verb as an independent token. Addressing this requires SVC-aware tokenization schemes and dictionaries that encode multi-verbal units explicitly. More broadly, punctuation marks and non-alphanumeric symbols should not be removed from preprocessing pipelines without prior linguistic analysis, as they may serve as tone or prosodic markers in several of these languages, and their removal can introduce irrecoverable information loss.

\textbf{\textsc{Regional Collaboration and Cross-Lingual Transfer:}} Given the linguistic continuity between Senegalese and neighboring West African languages, regional collaboration represents a major opportunity. Shared resources across Wolof, Pulaar, Mandingue, and Soninké communities could enable the development of multilingual models with stronger generalization, and thus reduce the gap between Wolof and the other local languages. Partnerships with initiatives such as \texttt{Masakhane}\footnote{\url{https://www.masakhane.io/}}, \texttt{EthioNLP}\footnote{\url{https://ethionlp.github.io/}}, \texttt{Mak-AI}\footnote{\url{https://air.ug/}} and \texttt{Sunbird AI}\footnote{\url{https://sunbird.ai/}} can enhance interoperability and ensure the inclusion of underrepresented languages in continental benchmarks. As an example, a historic milestone for African AI Research has been achieved for the first time, by a team from \texttt{\textsc{InstaDeep (Tunisia)}}\footnote{\url{https://instadeep.com/}} and \texttt{\textsc{Stellenbosch University (South Africa)}}, by earning an Oral presentation at \texttt{\textsc{NeurIPS}}\footnote{Neural Information Processing Systems.} \citep{chalumeau2025}, one of the most prestigious AI conference in the world \citep{google-scholar}. As of today, the continent contributes less than 3\% to the global AI market \citep{g20}, and these initiatives highlight the critical need to strengthen the key ingredient: \texttt{\textsc{Partnerships}}. Public-private partnerships and cross-border collaborations have therefore a huge potential to accelerate AI adoption and fuel a generation of African innovators.

\textbf{\textsc{Open Data, Tools, and Infrastructure:}} Expanding the availability of open data and reproducible tools is essential for building capacity and trust.
Many master's level projects and theses are also carried out on Senegalese languages, but remain difficult to gather due to their lack of visibility. The centralized GitHub repository introduced in this work serves as a living catalogue of Senegalese NLP resources, facilitating transparency and reuse. It aggregates datasets, models, and tools across all NLP tasks covered in this survey, with direct links to source repositories and, where available, license information. Institutional support and community contributions could further extend this infrastructure through GPU and open data grants, national cloud storage, and public data portals. Data collection should also be included in training curricula, as in this \texttt{AMMI} program\footnote{\url{https://github.com/besacier/AMMIcourse}}, in which each student (or duo of students) had to record 02 hours of speech in their native languages. The development of multilingual and multitask benchmarks such as \texttt{XTREME-UP} \citep{Ruder2023} is also important in order to facilitate the evaluation of models on these languages and foster their inclusion. Beyond data collection, ensuring data quality demands dedicated protocols tailored to these languages. Developing or extending word lists and dictionaries that reflect standardized orthographies, so that automated cleaning can distinguish plausible forms from noise. This will allow to systematically inspect sizeable samples of any multilingual web-crawled corpus before use, since large datasets such as \texttt{OSCAR} \citep{Ortiz2019} and \texttt{mC4} \citep{xue-etal-2021-mt5} have been shown to contain significant levels of mislabelled or erroneous content for low-resource African languages \citep{adebara-abdul-mageed-2022-towards}. 

\textbf{\textsc{Human Capital and Interdisciplinary Training:}} Bridging the skills gap requires long-term, and multi-layered investment in education and workforce development. \textit{Hybrid pathways}, i.e integrating formal education, project-based learning, online training, and industry apprenticeships, have been identified as one of the fastest route to skilled practitioners by 72.7\% of African AI ecosystem stakeholders \citep{gsma2026talent}. Embedding NLP and data science curricula within local universities, particularly in linguistics departments, can produce a new generation of computational linguists fluent in \texttt{both the technical and cultural dimensions of Senegalese languages}. This must be complemented by strong industry-academia partnerships, as weak institutional linkages represent the most frequently cited barrier to effective AI talent development (reported by 45.5\% of ecosystem respondents) \citep{gsma2026talent}. \texttt{Community-driven programmes} have therefore become essential complements to formal training pathways: the Deep Learning Indaba network\footnote{\url{https://deeplearningindaba.com/}} expanded from 18 affiliated countries in 2018 to approximately 47 by 2025, with annual conference attendance growing from roughly 200 to over 1,000 participants, driven in part by the increasing availability of specialised academic programmes \citep{gsma2026talent}. Specifically, Senegal hosted the 2024 edition thanks to \texttt{AI Hub Senegal}\footnote{\url{https://www.ai-hubsenegal.sn/}}  and the \texttt{GalsenAI} community, which has been organizing the local version (IndabaX) since 2018 \footnote{\url{https://galsen.ai/activites/indabax/}}. This underscores the importance and ability of communities of practitioners to promote the local ecosystem on the international stage.
\texttt{Open-source participation} in global developer communities, rated the most effective global learning pathway (4.5 out of 5.0 across the continent), further enables Senegalese NLP researchers to build practical skills and contribute to shared linguistic AI infrastructure.

\textbf{\textsc{Applied Research and Societal Impact:}} Beyond academic exploration, NLP technologies can yield tangible benefits for governance, education, and cultural preservation.  In the social sciences, language models can accelerate ethnographic analysis, making research outputs more inclusive and timely. \texttt{\textsc{CLAD}}\footnote{\url{https://clad.ucad.sn/}} and \texttt{\textsc{IFAN}}\footnote{\url{https://ifan.ucad.sn/}}, for example, have tremendous potential to launch the first \texttt{national applied research laboratory in NLP for Senegalese languages} and initiate projects such as \texttt{\textsc{NaijaVoices}}\footnote{https://naijavoices.com/} to equip our local languages. This will allow for the promotion of \textbf{computational social science} and modernize the research approaches of local social science researchers. One structural prerequisite that the community must address in parallel is the improvement of digital literacy in the national languages themselves. As \citet{adebara-abdul-mageed-2022-towards} observe, without a literate population in a language, the feedback loops that sustain NLP development; data collection, annotation, error correction, and community validation; cannot function properly. \textit{Without improvement of literacy in African languages, there is no flourishing future for African NLP.} Advocacy for language policies that extend the use of national languages in media, government, and education beyond the current \texttt{MOHEBS} elementary school program is therefore not peripheral to NLP research but a structural prerequisite for it. Concretely, a community of readers and writers in Wolof, Pulaar, and the remaining four national languages is the foundational ingredient from which training corpora, human evaluators, and domain-specific experts can be drawn at scale.

\textbf{\textsc{Toward a Sustainable and Ethical Ecosystem:}} Sustainability depends on continuous coordination among policymakers, researchers, and communities. Embedding ethical data governance into the national AI framework will help safeguard cultural heritage while promoting equitable innovation. Long-term public–private partnerships and regional alliances will be key to scaling these initiatives. The communities engaged in NLP resource development should also be actively extended beyond AI and machine learning to include theoretical linguists, anthropologists, sociologists, field workers, and other scholars with knowledge of these languages and their cultural contexts \citep{adebara-abdul-mageed-2022-towards}. This interdisciplinary broadening ensures that datasets reflect genuine linguistic diversity rather than the artifacts of a small, technically-trained contributor base, and that evaluation criteria capture the communicative realities that matter to speakers. Special vigilance is moreover required against predatory uses of data collected from local communities. Monitoring, censorship, and surveillance applications,  can cause direct harm to vulnerable populations and irreparably erode the community trust without which participatory data collection cannot succeed \citep{adebara-abdul-mageed-2022-towards}. Senegal thus has the opportunity to become a continental reference for inclusive, ethical, and locally grounded NLP development, provided that data sovereignty, speaker consent, and community benefit remain at the center of every initiative.

\section{Conclusion}
This paper has presented the first comprehensive synthesis of Natural Language Processing initiatives and resources for the six national languages of Senegal: \texttt{Wolof}, \texttt{Pulaar}, \texttt{Sérère}, \texttt{Diola}, \texttt{Mandingue}, and \texttt{Soninké}; while situating them within the broader context of the social sciences. We have shown that, although Senegal’s NLP ecosystem is still emerging, it is underpinned by strong institutional, academic, and community foundations. Current local and regional initiatives are nevertheless catalyzing a shift from isolated experiments to a coordinated, nationally anchored framework. The publicly available datasets, models, and tools are released in a centralized GitHub repository\footnote{\url{https://github.com/DerXter/State-of-NLP-Research-in-Senegal}}. The integration of NLP into the social sciences offers transformative potential: automating transcription, translation, and thematic analysis can significantly enhance research productivity and inclusiveness. Yet this potential will only be realized through sustained investment in open data, interdisciplinary training, and ethical governance. The establishment of shared infrastructures (computational, linguistic, and institutional) will ensure that NLP development in Senegal reflects both global best practices and local realities. Continued collaboration between linguists, computer scientists, policymakers, and local communities will be essential to build a fair and enduring digital future for all Senegalese languages.

The findings of this survey provide a foundation for future research on NLP for Senegalese languages. They offer researchers a means of identifying priority areas for further investigation and formulating research agendas that directly address the challenges and opportunities highlighted by the survey. In future work, we plan to explore a qualitative evaluation of existing datasets in Senegalese languages as well as prospects for implementing benchmarks in various NLP tasks. Platforms such as \texttt{\textsc{WhatsApp}} also offer unprecedented opportunities to conduct large-scale data collection campaigns as in \citep{ahesi} and could be the subject of further study.

\section*{Acknowledgements}
This project has been funded by \textsc{Fondation Botnar}.




\bibliographystyle{elsarticle-harv} 
\bibliography{example}






\end{document}